\documentclass[10pt,journal,compsoc]{IEEEtran}

\usepackage{cite}
\usepackage{graphicx}  
\usepackage{amsmath,amssymb,amsfonts}
\usepackage{multirow}
\usepackage{enumitem}
\usepackage{subcaption} 
\usepackage{float} 
\usepackage{algorithm}
\usepackage{algpseudocode}
\usepackage{color}
\usepackage{booktabs}
\usepackage{lineno}
\usepackage{url} 

\begin{document}
\title{Dynamic Graph Convolutional Recurrent Network for Traffic Prediction: Benchmark and Solution}
\author{Fuxian Li,
        Jie Feng,
        Huan Yan,
        Guangyin Jin,
        Depeng Jin,~\IEEEmembership{Senior Member,~IEEE}
        and Yong~Li,~\IEEEmembership{Senior Member,~IEEE}
\IEEEcompsocitemizethanks{\IEEEcompsocthanksitem F.~Li, J.~Feng, H.~Yan, D.~Jin, Y.~Li are with Beijing National Research Center for Information Science and Technology (BNRist), Department of Electronic Engineering, Tsinghua University, Beijing 100084, China.\protect\\
E-mail: lifx19@mails.tsinghua.edu.cn, fengj12ee@hotmail.com, yanhuanthu@gmail.com, \{jindp, liyong07\}@tsinghua.edu.cn
\protect\\
\IEEEcompsocthanksitem G.~Jin is with College of Systems Engineering, National University of Defense Technology, Changsha, China.\protect\\
E-mail: jinguangyin18@nudt.edu.cn.
}}

\markboth{IEEE TRANSACTIONS ON KNOWLEDGE AND DATA ENGINEERING}
{Shell \MakeLowercase{\textit{et al.}}: Bare Demo of IEEEtran.cls for Computer Society Journals}

\IEEEtitleabstractindextext{%
\begin{abstract}
Traffic prediction is the cornerstone of intelligent transportation system. Accurate traffic forecasting is essential for the applications of smart cities, i.e., intelligent traffic management and urban planning. 
Although various methods are proposed for spatio-temporal modeling, they ignore the dynamic characteristics of correlations among locations on road network. Meanwhile, most Recurrent Neural Network (RNN) based works are not efficient enough due to their recurrent operations. Additionally, there is a severe lack of fair comparison among different methods on the same datasets. 
To address the above challenges, in this paper, we propose a novel traffic prediction framework, named Dynamic Graph Convolutional Recurrent Network (DGCRN). In DGCRN, hyper-networks are designed to leverage and extract dynamic characteristics from node attributes, while the parameters of dynamic filters are generated at each time step. 
We filter the node embeddings and then use them to generate dynamic graph, which is integrated with pre-defined static graph.
As far as we know, we are first to employ a generation method to model fine topology of dynamic graph at each time step. 
Further, to enhance efficiency and performance, we employ a training strategy for DGCRN by restricting the iteration number of decoder during forward and backward propagation.
Finally, a reproducible standardized benchmark and a brand new representative traffic dataset are opened for fair comparison and further research. Extensive experiments on three datasets demonstrate that our model outperforms 15 baselines consistently. Source codes are available at \url{https://github.com/tsinghua-fib-lab/Traffic-Benchmark}.

\end{abstract}

}

\maketitle

\IEEEdisplaynontitleabstractindextext

\IEEEpeerreviewmaketitle


\section{Introduction}
With the rapid development of urbanization, transportation system in the city is under great pressure when facing growing populations and vehicles. Fortunately, advances in data intelligence and urban computing have made it possible to collect massive amounts of traffic data and conduct analysis. The traffic data can be crowd flow data or traffic speed data, which not only is an important indicator reflecting the state of transportation system, but also is used for prediction tasks of future traffic conditions. 
Among these tasks, traffic prediction acts as the cornerstone of intelligent transportation system (ITS) with the aim of understanding and developing an optimal transportation system and minimal traffic congestions. By predicting future traffic, 
it provides reference for urban planning and traffic management, so as to reduce congestion and improve traffic efficiency, and provide early warning for public safety emergency management. Accurate traffic prediction can also help travelers plan and change their routes, thus improving their quality of life.

Traffic prediction originates from single variant time series forecasting with many traditional statistic-based methods, e.g., history average (HA), vector auto-regression (VAR), support vector regression (SVR) and auto-regressive integrated moving average (ARIMA). However, most of these methods need to satisfy the stationarity assumption for each time series. Besides, there are not many parameters in these methods, most of the which need to be designed by experts rather than mined from data. Thus, traditional methods are restricted by their ability for capturing spatio-temporal dependency.

In deep learning, temporal correlations can be captured by RNN, Convolutional Neural Network (CNN) or attention. Meanwhile, spatial dependencies can be modeled by CNN, Graph Neural Network (GNN) or attention.
At first, grid-based graph is used to build spatial topology and then fed into CNNs~\cite{stresnet, deepstn+, dmvst, stdn}, but fail to deal with non-euclidean data. The appearance of graph neural network makes the deep learning model able to tackle non-euclidean data and the road distance between nodes (sensors or road segments) are often used to calculate the weights of edges~\cite{dcrnn, stgcn, stmetanet, mrabgcn, gman, stgrat, sttns, gwn, stgnn}. Besides, the self-adaptive adjacency matrix is proposed to preserve hidden spatial dependencies~\cite{gwn, mtgnn, agcrn, stgnn, slcnn}. However, it is tough for the static graph to adjust itself dynamically and capture complex dynamic spatial dependencies. CNN can also be used to capture temporal correlations~\cite{stgcn, astgcn, stsgcn, lsgcn, gwn, mtgnn} efficiently at the expense of flexibility due to its implicit temporal modeling, which makes the time steps invisible. Self-attention is effective to capture global dependency~\cite{gman} but not good at short-range prediction. RNN is powerful to capture sequential correlations with relatively higher time consuming~\cite{agcseq2seq, tgclstm, stmgcn, stmetanet, dgcn, stgnn, dcrnn,mrabgcn,agcrn}.

Since traffic data shows strong dynamic correlations in the spatial and temporal dimension, 
it is crucial to model dynamic and nonlinear spatio-temporal correlations for accurate traffic prediction.
The spatial correlations between different locations are highly dynamic, which is determined by real-time traffic conditions and the topology of traffic network. Meanwhile, the temporal dependencies are also complex due to the mixture of non-linear changes and periodicity.

Due to the complex spatio-temporal correlations, 
we need to address the following challenges:
\begin{itemize}
    \item The dynamic characteristics of traffic graph topology need to be further modeled. The pre-defined adjacency matrix and adaptive adjacency matrix are both static with the change of time, which is not enough to reflect the dynamic characteristics of road network topology in reality.
    \item Static distance based graph and dynamic attribute based graph describe the topology of traffic network from distinct perspectives, thus integrating them can give model a wider range of horizon for capturing spatial dependencies. 
    However, most works fail to fuse them together while maintaining efficiency and avoiding over-smoothing.
    \item Although widely used in time series prediction, the training speed of RNN and its variants are restricted by the inner recurrent operations, which blocks the applications of architectures like sequence-to-sequence in traffic prediction task.
    \item The number of proposed models and traffic datasets is increasing with rapid growing of traffic prediction field, where the models are evaluated on different datasets with corresponding experimental settings. This makes it tricky to make a fair comparison among models, impeding the development of the field.
    
\end{itemize}

In this work, we propose a hyper-network to generate dynamic adjacency matrix step by step adaptively, where the message passing of dynamic node features improve the effectiveness of graph generation. Then we integrate the static graph and dynamic graph together in graph convolution module, which improve the performance significantly. To improve the efficiency of the seq2seq architecture, we employ a general training strategy for RNNs. For more challenging prediction task, we publish a representative congestion-related traffic dataset. Finally, We propose an open-source benchemark based on three public datasets and 15 baselines for fair comparison and further research.

The main contributions of our paper are summarized as follows:
\begin{itemize}
    \item We propose a GNN and RNN based model, where the dynamic adjacency matrix is designed to be generated from a hyper-network step by step synchronize with the iteration of RNN. The dynamic graph is incorporated with the pre-defined graph and skip connection to describe the dynamic characteristics of road networks more effectively, enhancing the performance of prediction.
    \item We employ a general training method for RNN based models, which not only improves performance effectively and efficiently, but also sharply decreases training time consumption, in order to overcome the shortcomings of RNNs in efficiency and resource occupancy.
    \item We conduct extensive experiments based on three publicly available datasets to demonstrate the effectiveness of our proposed model on the traffic prediction task. Compared with 15 baselines in benchmark, results demonstrate that our model can reduce the error of prediction significantly and achieve state-of-the-art prediction accuracy.
    \item We publish a new real-life congestion-related dataset\footnote{The source code and dataset are available at \url{https://github.com/tsinghua-fib-lab/Traffic-Benchmark}.} and conduct sufficient comparative experiments on three public datasets to get fair comparison results for various complex models and derive a reproducible standardized benchmark. The benchmark is composed of comparative prediction performance of 15 representative methods on 3 datasets, which can provide reference value for researchers to facilitate further research on related problems.
\end{itemize}

The rest of this paper is organized as follows. We first review the related works and propose a comprehensive classification in Section 2. Then the traffic prediction problem is formulated in Section 3. Motivated by the challenges, we introduce the details of our solutions in Section 4. After that, we evaluate our model by three real-world traffic datasets and derive the traffic benchmark in Section 5, where the ablation studies and parameter studies are conducted. Finally, we conclude our paper in Section 6. 
\section{Literature Review}
\begin{table*}[!h]
\centering
\caption{Model classification.}
\label{tab:classification-table}
\setlength{\tabcolsep}{0.7mm}{
\begin{tabular}{|c|c|c|c|c|}
\hline
\textbf{Model} & \textbf{Spatial   topology construction} & \textbf{Spatial dependency } & \textbf{ Temporal dependency } & \textbf{External features} \\ \hline
\textbf{DCRNN~\cite{dcrnn}} & Distance-based graph & GCN & GRU & Time \\ 
AGC-Seq2Seq~\cite{agcseq2seq} & Binary graph & GCN & GRU + Attention & None \\ 
TGC-LSTM~\cite{tgclstm} & Binary graph + Others & GCN & LSTM & None \\ 
ST-MGCN~\cite{stmgcn} & Binary graph + Others & GCN & RNN & POI + Road network structure \\ 
\textbf{ST-MetaNet~\cite{stmetanet}} & Distance-based graph & Meta-GAT & Meta-GRU & Time + Road network structure \\ 
MRA-BGCN~\cite{mrabgcn} & Distance-based graph + Others & GCN + Attention & GRU & Time \\ 
DGCN~\cite{dgcn} & Binary graph & GCN + Attention & LSTM + Attention & Time \\ 
\textbf{AGCRN~\cite{agcrn}} & Adaptive graph & GCN & GRU & Time \\ 
STGNN~\cite{stgnn} & Adaptive graph + Distance-based graph & GCN & GRU + Transformer & Time \\ \hline
\textbf{GMAN~\cite{gman}} & Distance-based graph & Graph embedding +   Attention & Embedding + Attention & Time \\ 
ST-GRAT~\cite{stgrat} & Distance-based graph & Graph embedding +   Attention & Embedding + Attention & Time \\ 
STTNs~\cite{sttns} & Distance-based graph & Transformer & Transformer & Time \\ \hline

\textbf{STGCN~\cite{stgcn}} & Distance-based graph & GCN & CNN & Time \\ 
\textbf{ASTGCN~\cite{astgcn}} & Binary graph & GCN + Attention & CNN + Attention & Time \\ 
\textbf{STSGCN~\cite{stsgcn}} & Binary graph & GCN & GCN & Time \\ 
LSGCN~\cite{lsgcn} & Binary graph & GCN + Attention & CNN & Time \\ 
\textbf{Graph WaveNet~\cite{gwn}} & Adaptive graph + Distance-based graph & GCN & CNN & Time \\
SLCNN~\cite{slcnn} & Adaptive graph & GCN & CNN & Time \\ 
\textbf{MTGNN~\cite{mtgnn}} & Adaptive graph & GCN & CNN & Time \\ \hline

ST-ResNet~\cite{stresnet} & Grid-based graph & CNN & CNN & Time + Weather \\ 
DeepSTN+~\cite{deepstn+} & Grid-based graph & CNN & CNN & Time + POI \\ 
DMVST-Net~\cite{dmvst} & Grid-based graph + Others & CNN + Graph embedding & RNN & Time + Weather \\ 
STDN~\cite{stdn} & Grid-based graph & CNN & RNN + Attention & Time \\ \hline
DARNN~\cite{darnn} & None & None & RNN + Attention & None \\ 
GeoMAN~\cite{geoman} & None & Attention & RNN + Attention & Time + Weather + POI  \\ \hline
\end{tabular}}
\end{table*}

In traffic prediction, utilizing prior geo-information is of great importance. Since grid-based data~\cite{stresnet, deepstn+, dmvst, stdn} has natural limit on representing complex spatial topology, non-Euclidean pre-defined graphs are introduced, including distance-based graph~\cite{dcrnn, stgcn, stmetanet, mrabgcn, gman, stgrat, sttns, gwn, stgnn}, binary graph~\cite{agcseq2seq, tgclstm, stmgcn, astgcn, stsgcn, dgcn, lsgcn}, multi-view graphs~\cite{stmgcn} and adaptive graphs~\cite{gwn, mtgnn, agcrn, stgnn, slcnn}, etc. 
Besides the aforementioned spatial topology construction, spatio-temporal dependency modeling is also core of traffic prediction.
CNN~\cite{stresnet, deepstn+, dmvst, stdn}, GNN~\cite{dcrnn, stgcn, stmetanet, mrabgcn, dgcn, gwn, mtgnn, agcrn, slcnn, stgnn, agcseq2seq, tgclstm, stmgcn, astgcn, stsgcn, lsgcn} and attention mechanism~\cite{mrabgcn, stmetanet, dgcn, gman, stgrat, sttns, astgcn, lsgcn, geoman} have been widely used to capture spatial dependency. Meanwhile, most works employ RNNs~\cite{agcseq2seq, tgclstm, stmgcn, stmetanet, dgcn, stgnn, dcrnn,mrabgcn,agcrn}, TCNs~\cite{stgcn, astgcn, stsgcn, lsgcn, gwn, mtgnn} and attention~\cite{agcseq2seq, dgcn, stgnn, gman, stgrat, sttns, astgcn, stdn, darnn, geoman} to model temporal dependency.

Additionally, as traffic conditions are highly susceptible to external environment, most traffic prediction works introduce external features such as weather~\cite{stresnet}, Point of Interest (POI)~\cite{deepstn+, stmgcn}, time of day~\cite{dcrnn, stgcn, stmetanet, mrabgcn, dgcn, gman, stgrat, sttns, gwn, stgnn, stresnet, deepstn+, dmvst, stdn} and day of week~\cite{stresnet, deepstn+} into the model to enhance performance. 

Based on the above analysis, we classify the related works based on four parts: spatial topology construction, spatial dependency modeling, temporal dependency modeling and external features, as shown in Table~\ref{tab:classification-table}.
\begin{itemize}
    \item \textbf{Spatial topology construction.} Traditional methods have no need to construct the spatial topology in advance. Convolutional neural network needs grid-based graph, where the spatial topology construction method is to partition a map into H$\times$W equal size grids, where H, W represent the height and width of the grid-based map respectively~\cite{stresnet}\cite{deepstn+}\cite{dmvst}\cite{stdn}\cite{stmetanet}. Further, graph neural networks break the restrictions and can be used for non-Euclidean data which is more general and flexible to describe the real-world road networks. The original graph construction method is to calculate the similarity between pairs of nodes by a distance metric with thresholded Gaussian kernel function and get a weighted adjacency matrix~\cite{dcrnn, stgcn, stmetanet, mrabgcn, gman, stgrat, sttns, gwn, stgnn} or just use the connectivity to derive a simple binary adjacency matrix~\cite{agcseq2seq, tgclstm, stmgcn, astgcn, stsgcn, dgcn, lsgcn}. Some works also build graphs from multiple views, like POI similarity~\cite{stmgcn}, DTW~\cite{dtw} similarity~\cite{dmvst}, free-flow reachable matrix~\cite{tgclstm}, edge-wise graph~\cite{mrabgcn}, etc. However, the pre-defined adjacency matrix is static and the construction methods are limited, which is a natural shortcoming on describing the complex road networks. To make up for the ``lost view", the adaptive adjacency matrix~\cite{gwn} is proposed and achieve improvement on performance, where the parameters of the adjacency matrix are learnable with training process. Specifically, some works directly make the adjacency matrix learnable~\cite{slcnn}, others calculate similarities among learnable node embeddings~\cite{gwn, mtgnn, agcrn, stgnn}. However, it is still hard for static adaptive adjacency matrix to model the dynamic characteristics of road networks.~\cite{stsgcn} connects all nodes with themselves at the adjacent time steps to get a localized spatial-temporal graph from which the correlations between each node and its spatial-temporal neighbors can be captured directly. However, this method only considers the nearest time steps and the representation ability is limited.

    \item \textbf{Spatial dependency modeling.} In early period, traffic prediction is seen as a simple time series prediction task. Most traditional methods, like auto-regressive (AR), moving average (MA), auto-regressive moving average (ARMA) and auto-regressive integrated moving average (ARIMA~\cite{box1970time}), only focus on time series of single variant and fail to capture the correlations among variants, which also severely restrict the efficiency when facing time series of multiple variants. What's more, most traditional statistical methods need strong stationary-related assumptions which leads to extreme failure on predicting time series with high fluctuation and many missing values. In recent years, With the development of deep learning, convolutional neural network is widely used in capturing spatial correlations from Euclidean data ~\cite{stresnet,deepstn+,dmvst,stdn,stmetanet}. Graph embedding can capture spatial dependency and generate encoded embedding vectors for subsequent processing. \cite{gman,stgrat,dmvst} use graph embedding to derive the inputs of the following modules. Graph neural network is powerful and flexible to capture spatial correlations in non-Euclidean data. \cite{mtgnn,stgnn,agcrn,mrabgcn,slcnn,dcrnn,gwn} utilize graph convolutional network (GCN~\cite{Kipf2016Semi}) and its variants. \cite{stmetanet} uses meta learning to generate the parameters of the graph attention network (GAT~\cite{gat}). \cite{gman,stgrat,astgcn,sttns,geoman,lsgcn} design various architectures of spatial attention to capture spatial correlations.~\cite{stsgcn} propose spatial-temporal synchronous Graph Convolution Module to capture spatial correlations and temporal correlations simultaneously.

    \item \textbf{Temporal dependency modeling.} RNN and its variants are proposed for sequential data and are naturally proper and also powerful for capturing temporal dependency\cite{tgclstm,stgnn,stmgcn,darnn,geoman,dmvst,stdn,agcseq2seq}. The recurrent operation of RNNs leads to flexibility of model architecture at the cost of time and memory consuming. The presentation of GRU~\cite{gru} and LSTM~\cite{lstm} had further enhanced RNNs' ability of modeling long-range sequential dependency and avoid the gradient vanishing. \cite{dcrnn,mrabgcn} replace the fully connected neural network in GRU~\cite{gru} with graph neural network which is effective to integrate message passing with recurrent operation and gate mechanism. CNN~\cite{cnn} has been proved to be effective and efficient to capture temporal correlations in sequential data~\cite{mtgnn,slcnn,lsgcn,gwn,astgcn,stgcn} but the architecture is relatively more solid and there is little room for adjustment to adapt different situations. Attention mechanism is also powerful and helpful in sequential dependency modeling, including self-attention~\cite{stgnn,gman,stgrat,sttns} and other variants~\cite{darnn,geoman,stdn,agcseq2seq}. Represented by Transformer~\cite{transformer}, self-attention based models have governed the NLP area. Benefitting from the ability of capturing global sequential dependency and parallelization, self-attention based models are still restricted by large memory consuming and high demand for computation source. Besides, the positional encoding is simple but not enough to represent the complex sequential information while the self-attention can only derive the weights of importance, thus capturing local (short-term) sequential dependency is harder for self-attention based models than RNNs.
    
    \item \textbf{External features.} It is obvious that traffic condition can be significantly affected or inferred by external factors such as weather~\cite{stresnet,dmvst} and POI~\cite{deepstn+,stmgcn}, which need extra collection and preprocessing. Some other features, like time information~\cite{stresnet, deepstn+, dcrnn, gwn, mtgnn}, DTW similarity~\cite{dmvst}, road structure and connectivity~\cite{stmetanet, stmgcn}, can be generated from original time series or adjacency matrix. Besides, making different external factors cooperate well with each other and even generate new useful features is critical and promising for model designers.
\end{itemize}

Compared with the aforementioned works, our model employs graph generation at each time step to carry out dynamic and fine-grained modeling of road network topology. We also fuse the generated dynamic graph with the static graph effectively. Besides, the employed training strategy for RNNs improves both efficiency and performance significantly.

\section{Preliminaries and Problem Formulation}
Traffic prediction task can be formulated as a multivariate time series forecasting problem with auxiliary prior knowledge. 
Generally, the prior knowledge is the pre-defined adjacency matrix denoted as a weighted directed graph $\mathcal{G}=(\mathcal{V},\mathcal{E},A)$. Here, $\mathcal{V}$ is a set of $N = |\mathcal{V}|$ vertices which represent different locations (e.g., traffic sensors or road segments) on the road network; $\mathcal{E}$ is a set of edges and $A \in \mathbb{R}^{N \times N}$ is the weighted adjacency matrix where each element represents the proximity between vertices measured from certain view (e.g. road network distance, DTW similarity, POI similarity, etc.).

The multivariate time series, aka the graph signals, can be denoted as a feature tensor $\mathbf{X} \in \mathbf{R}^{T\times N\times D}$ on graph $ \mathcal{G} $, where $T$ is the length of sequence and $ D $ is the number of features of each node (e.g., traffic volume, traffic speed, etc.). At each time step $ t $, the graph signal is $\mathbf{X}_t \in \mathbf{R}^{N\times D}$. Finally, the traffic prediction problem can be formulated as follows:

\textbf{Problem Formulation}. Given the graph $ \mathcal{G}=(\mathcal{V},\mathcal{E},A) $ and its observed $P$ step graph signals $\mathbf{X}_{(t-P):t}$, to learn a function $f$ which is able to map $\mathbf{X}_{(t-P):t}$ and $ \mathcal{G} $ to next $Q$ step graph signals $ \hat{\mathbf{X}}_{t:(t+Q)}$, represented as follows:
\begin{equation}
    [\mathbf{X}_{(t-P):t},\mathcal{G}] \xrightarrow[]{f} \hat{\mathbf{X}}_{t:(t+Q)}, 
\end{equation}
where $\mathbf{X}_{(t-P):t} = (\mathbf{X}_{t-P},\mathbf{X}_{t-P+1},...,\mathbf{X}_{t-1}) \in \mathbf{R}^{P \times N \times D}$ and  $\hat{\mathbf{X}}_{t:(t+Q)} = (\hat{\mathbf{X}}_{t},\hat{\mathbf{X}}_{t+1},...,\hat{\mathbf{X}}_{t+Q-1}) \in \mathbf{R}^{Q \times N \times D}$.
\section{Methodolody}

\begin{figure*}[t]
	\centering
    \includegraphics[width=1.0\textwidth]{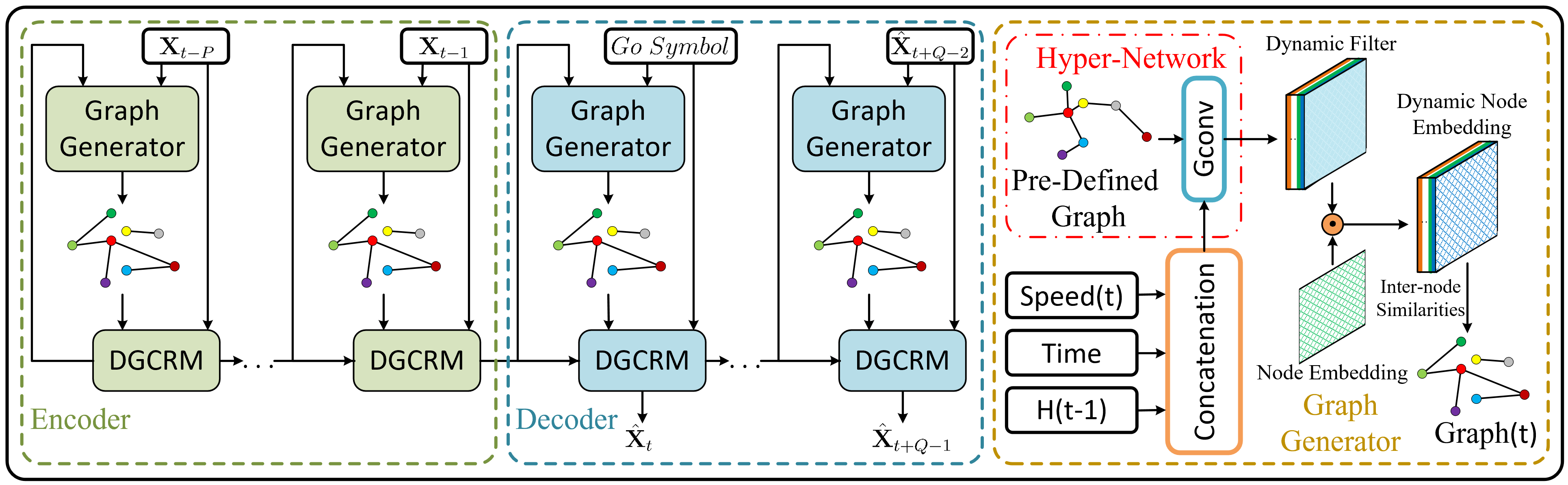}	
	\caption{The architecture of DGCRN.}
	\label{fig:framework}
\end{figure*}
\begin{figure}[t]
	\centering
	\includegraphics[width=0.49\textwidth]{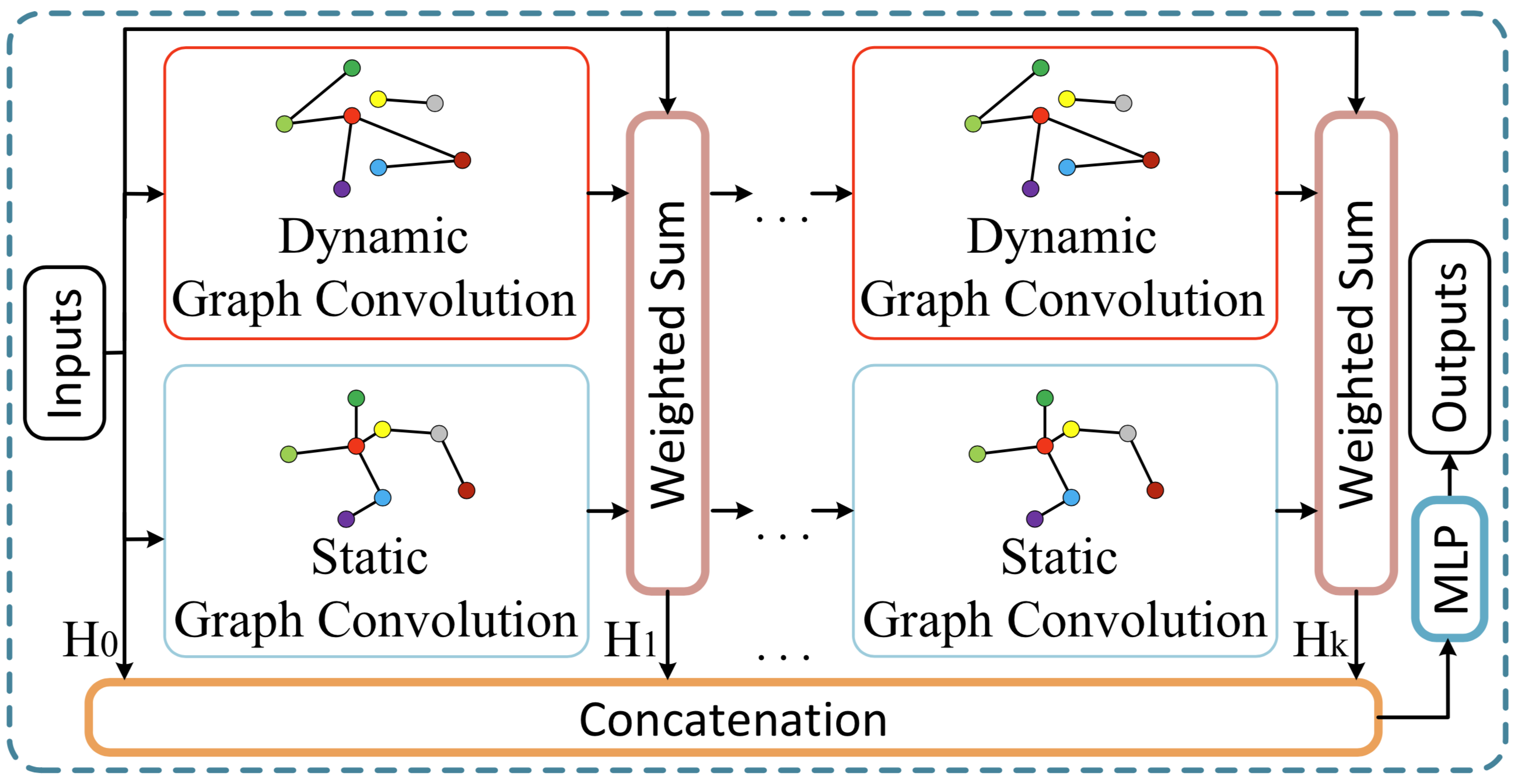}	
	\caption{Structure of dynamic graph convolution module.}
	\label{fig:gcn_new}
\end{figure}

The recurrent operations of RNNs bring about dynamic knowledge which is, however, not fully utilized for capturing dynamic spatio-temporal correlations. 
Following this idea, we design the DGCRN based on a sequence-to-sequence architecture~\cite{seq2seq} including an encoder and a decoder, as shown in Figure~\ref{fig:framework}. 
The dynamic adjacency matrix at each time step is generated synchronize with the recurrent operation of DGCRN where the two graph generators are designed for encoder and decoder respectively. After that, both the generated dynamic graph and the pre-defined static graph are used for graph convolution. Then, we use weighted sum to fuse the result of different graph convolution with skip connection at each layer, as demonstrated in Figure~\ref{fig:gcn_new}. Finally, we replace all the fully connected layers in traditional Gated Recurrent Unit (GRU) with the dynamic graph convolution module to derive our Dynamic Graph Convolutional Recurrent Module (DGCRM).

Specifically, DGCRN consists of two main components:

\begin{itemize}
    \item \textbf{Graph generator.} To model the dynamic characteristics of road network topology, we design a novel graph generator. Specifically, we employ two hyper-networks for encoder and decoder to capture dynamic information and generate a dynamic filter, which is then combined with randomly initiated node embedding vectors and generate dynamic node embeddings. We finally calculate the pair-wise similarities between dynamic node embeddings and derive the dynamic adjacency matrix. 

    \item \textbf{Dynamic graph convolutional recurrent module.} To integrate the dynamic graph with static graph efficiently and effectively, we design a dynamic graph convolution module as shown in Figure~\ref{fig:gcn_new}. We use it to replace the fully connected layers in classical GRU to derive DGCRM which can capture temporal correlations as well as spatial correlations simultaneously. The dynamic adjacency matrix is incorporated with the static pre-defined adjacency matrix in graph convolution.

\end{itemize}

\subsection{Graph Generator}
The urban traffic conditions are complex and affected by diverse spatio-temporal correlations. It is natural to describe the traffic network from a dynamic view. We design a novel hyper-network for dynamic graph generation. The core of DGCRN is the update of the dynamic adjacency matrix $\mathbf{DA}^t$ at time step $t$ based on current and historical information. At each time step, the speed $\mathbf{V}_t$, time of day $\mathbf{T}_t$ and hidden state $\mathbf{H}_{t-1}$ are concatenated as the input of hyper-network:
\begin{align}
\mathbf{I}_t &= \mathbf{V}_t \parallel \mathbf{T}_t \parallel \mathbf{H}_{t-1},
\end{align}
where $\mathbf{I}_t \in \mathbf{R}^{B\times N\times D_{in}}$, $D_{in}$ is the feature dimension, $B$ is the batch size, $N$ is the number of nodes, $\parallel$ represents the concatenation operation. $\mathbf{I}_t$ is seen as dynamic node feature and then fed into graph convolution module: 
\begin{align}
\mathbf{DF}^t &= {\Theta}_{{\star}{G}} (\mathbf{I}_t),
\end{align}
where ${\Theta}_{{\star}{G}}$ represents the graph convolution and ${\Theta}$ denotes the learnable parameters. The pre-defined adjacency matrix $\mathbf{A}$ contains the static distance-based relations among nodes, which can be employed to conduct the message-passing process for dynamic node status. 
We call the above architectures the hyper-network whose output is a dynamic filter tensor $\mathbf{DF}^t \in \mathbf{R}^{B\times N\times D}$. We employ a position-wise multiplication operation between $\mathbf{DF}_t$ and the static randomly initiated node embedding ${\mathbf{E}} \in \mathbf{R}^{N \times D}$ whose parameters are learnable, where the broadcasting operation is required.

In practice, we employ two hyper-networks to generate two dynamic filters for source node embedding and target node embedding respectively, denoted as follows:
\begin{align}
\begin{split}
\label{eq:dgg}
\mathbf{DE}^t_1 &= tanh(\alpha (\mathbf{DF}^t_1 \odot \mathbf{E}_1),\\
\mathbf{DE}^t_2 &= tanh(\alpha (\mathbf{DF}^t_2 \odot \mathbf{E}_2),
\end{split}
\end{align}
where $\odot$ denotes the Hadamard product.
We employ similarity among nodes to calculate the dynamic adjacency matrix, following~\cite{mtgnn}:
\begin{equation}
\mathbf{DA}^t = ReLU(tanh(\alpha (\mathbf{DE}^t_1{\mathbf{DE}^t_2}^T-\mathbf{DE}^t_2{\mathbf{DE}^t_1}^T))),
\end{equation}
where $\alpha$ is a hyper-parameter to control the saturation rate of the activation function and $\mathbf{DA}^t \in \mathbf{R}^{B\times N\times N}$ denotes the dynamic adjacency matrix at time step $t$.

To sum up, by adding the hidden state of RNN to the input of graph generator, we integrate the iteration of RNN and the generation of dynamic graph simultaneously and seamlessly. Meanwhile, 
the graph generation becomes more effective owing to the graph convolution inside the hyper-network since it
applies the message passing among nodes and thus dynamic information can be fully utilized. 
As an essential part of graph convolution, the dynamic adjacency matrix generation module can learn a dynamic representation of road network implicitly, which can provide an effective complement for static distance-based adjacency matrix.

\subsection{Dynamic Graph Convolutional Recurrent Module}
The static distance-based graph and the dynamic node attribute based graph reflect the internode correlations from distinct perspectives.
To give model a wider range of horizon for traffic network, we combine dynamic graph with pre-defined graph when deploying graph convolutions, thus enhancing the performance of traffic prediction.
Specifically, we employ a weighted average sum of the graph convolution results of input graph signal, pre-defined static graph $\mathbf{A}$ and the dynamic graph $\mathbf{DA}^t$ at each graph convolution layer.

The k-hop dynamic graph convolution module can be denoted as follows:

\begin{align}
\begin{split}
\label{eq:gcn}
& \mathbf{H}^{(k)}=\alpha\mathbf{H}_{in} + \beta\mathbf{H}^{(k-1)}\Tilde{\mathbf{DA}^t} + \gamma\mathbf{H}^{(k-1)}\Tilde{\mathbf{A}}, \\
& \mathbf{H}_{out} = \sum_{i=0}^{K}\mathbf{H}^{(k)}\mathbf{W}^{(k)}, \mathbf{H}^{(0)}=\mathbf{H}_{in}, \\
& \Tilde{\mathbf{DA}^t} =\Tilde{\mathbf{D}^t}^{-1}(\mathbf{DA}^t+\mathbf{I}), \\
& \Tilde{\mathbf{D}}^t_{:,i,i}=1+\sum_j\mathbf{DA}^t_{:,i,j}, \\
& \Tilde{\mathbf{A}}=\Tilde{\mathbf{D}}^{-1}\mathbf{A}, \Tilde{\mathbf{D}}_{i,i}=\sum_j\mathbf{A}_{i,j},
\end{split}
\end{align}
where $\alpha$, $\beta$ and $\gamma$ are hyper-parameters to control the weights of different components. $\mathbf{H}_{in}$ and $\mathbf{H}_{out} = \Theta_{{\star}{G}} (\mathbf{H}_{in}, \mathbf{DA}^t, \mathbf{A})$ represent the input and output node states respectively.
$\mathbf{W}\in \mathbf{R}^{K\times D_{in}\times D_{out}}$ are learnable parameters, $K$ is the depth of propagation. Specifically, we adopt dual directional graph convolution to make good use of directed graphs, denoted as follows:
\begin{align}
\begin{split}
\label{eq:gcn_out}
& \mathbf{H}_{o} = {\Theta_1}_{{\star}{G}} (\mathbf{H}_{in}, \mathbf{DA}^t, \mathbf{A}) +  {\Theta_2}_{{\star}{G}} (\mathbf{H}_{in}, {\mathbf{DA}^t}^T, \mathbf{A}^T), 
\end{split}
\end{align}
which is shortened to $\mathbf{H}_{o} = \Theta_{{\star}{G}} (\mathbf{H}_{in})$ for brief.
When used as a part of the aforementioned hyper-network, the dynamic graph convolution module can be simplified by setting $\beta$ to 0, where the dynamic graph convolution is eliminated. 

RNN is powerful in modeling sequential dependency and GRU further improve RNN's ability on long-range temporal modeling and avoid gradient vanishing. Following~\cite{seo2016structured,dcrnn,mrabgcn}, we replace the matrix multiplications in GRU with dynamic graph convolution modules and get DGCRM, denoted as follows:
\begin{align} \label{eqn:gru_graph}
\begin{split}
	{z}^{(t)} &= \sigma({\Theta_z}_{{\star}{G}} (\mathbf{X}_{t} \parallel \mathbf{H}_{t-1})), \\
	{r}^{(t)} &= \sigma({\Theta_r}_{{\star}{G}} (\mathbf{X}_{t} \parallel \mathbf{H}_{t-1})), \\
	\tilde{{h}^{(t)}} &= \tanh({\Theta_h}_{{\star}{G}} (\mathbf{X}_{t} \parallel ({r}^{(t)} \odot \mathbf{H}_{t-1}))), \\
	\mathbf{H}_{t} &= {z}^{(t)} \odot \mathbf{H}_{t-1} + (1-{z}^{(t)}) \odot \tilde{{h}^{(t)}},
\end{split}
\end{align}
where $\mathbf{X}_{t}, \mathbf{H}_{t}$ denote the speed concatenated with time of day and output hidden state of DGCRM at time $t$ respectively, $ \odot $ represents the Hadamard product, $ \sigma ( \cdot ) $ denotes the sigmoid activation, ${r}^{(t)}, {z}^{(t)}$ are reset gate and update gate at time $t$, respectively.
${\star}{G}$ denotes the \textit{dynamic graph convolution} defined in Equation~\ref{eq:gcn} and $\Theta_z, \Theta_r, \Theta_h$ are learnable parameters for the corresponding graph convolution modules.

\subsection{Training Strategy}

\begin{algorithm}[t]
\caption{Training algorithm of DGCRN.}\label{alg}
\begin{algorithmic}[1]
\small
\State \textbf{Input}: Pre-defined graph $ \mathcal{G}=(\mathcal{V},\mathcal{E},A) $, graph signal tensor $\mathbf{X} \in \mathbf{R}^{T\times N\times D}$, initialized encoder's recurrent function $f_{en}(\cdot)$ with learnable parameters $\Theta_{en}$, initialized decoder's recurrent function $f_{de}(\cdot)$ with learnable parameters $\Theta_{de}$, learning rate $\gamma$, step size $s$, pre-defined function for scheduled sampling $f_{ss}(\cdot)$, zero tensor $\hat{\mathcal{Y}}\in R^{B\times Q\times N}$.
\State set $iter=1,i=1$
\Repeat
\State initialize hidden state $\mathbf{H}_0$, randomly select a batch (input $\mathcal{X}\in R^{B\times P\times N \times D}$, lable $\mathcal{Y} \in R^{B\times Q\times N}$, time of day $\mathcal{T} \in R^{B\times Q\times N} )$ from $\mathbf{X}$.
\If{$iter\%s==0$ and $i<Q$}
\State $i=i+1$
\EndIf

\For{$p$ in $0, 1,..., P-1$}
\State compute $\mathbf{H}_{p+1} = f_{en}(\mathcal{X}[: , p , :, :], \mathbf{H}_p, \mathcal{G};\Theta_{en})$
\EndFor

\State initialize graph signal $\mathcal{Y}_{in} \in R^{B\times N}$ as a zero tensor for decoder.
\For{$q$ in $0, 1,..., i-1$}
\State compute $\mathcal{Y}_{in} = \mathcal{Y}_{in} || \mathcal{T}[: , q, :]$
\State compute $\hat{\mathcal{Y}}[:, q, :] = f_{de}(\mathcal{Y}_{in}, \mathbf{H}_{P+q}, \mathcal{G};\Theta_{de})$
\State randomly select a number $c \sim \mathcal{U}(0,1)$.
\If{$c < f_{ss}(iter)$}
\State $\mathcal{Y}_{in}=\mathcal{Y}[: , q , :]$
\Else  
\State $\mathcal{Y}_{in}= \hat{\mathcal{Y}}[:, q, :]$
\EndIf
\EndFor

\State compute  $L=loss(\hat{\mathcal{Y}}[: , :i , :],\mathcal{Y}[: , :i , :])$
\State compute the stochastic gradient of $\Theta_{en}$ and $\Theta_{de}$ according to $L$.
\State update model parameters $\Theta_{en}$ and $\Theta_{de}$ according to their gradients and the learning rate $\gamma$.
\State $iter=iter+1$.
\Until{stopping criteria is met.}
\State output learned model.
\end{algorithmic}
\end{algorithm}

\begin{figure}[t]
	\centering
	\includegraphics[width=0.45\textwidth]{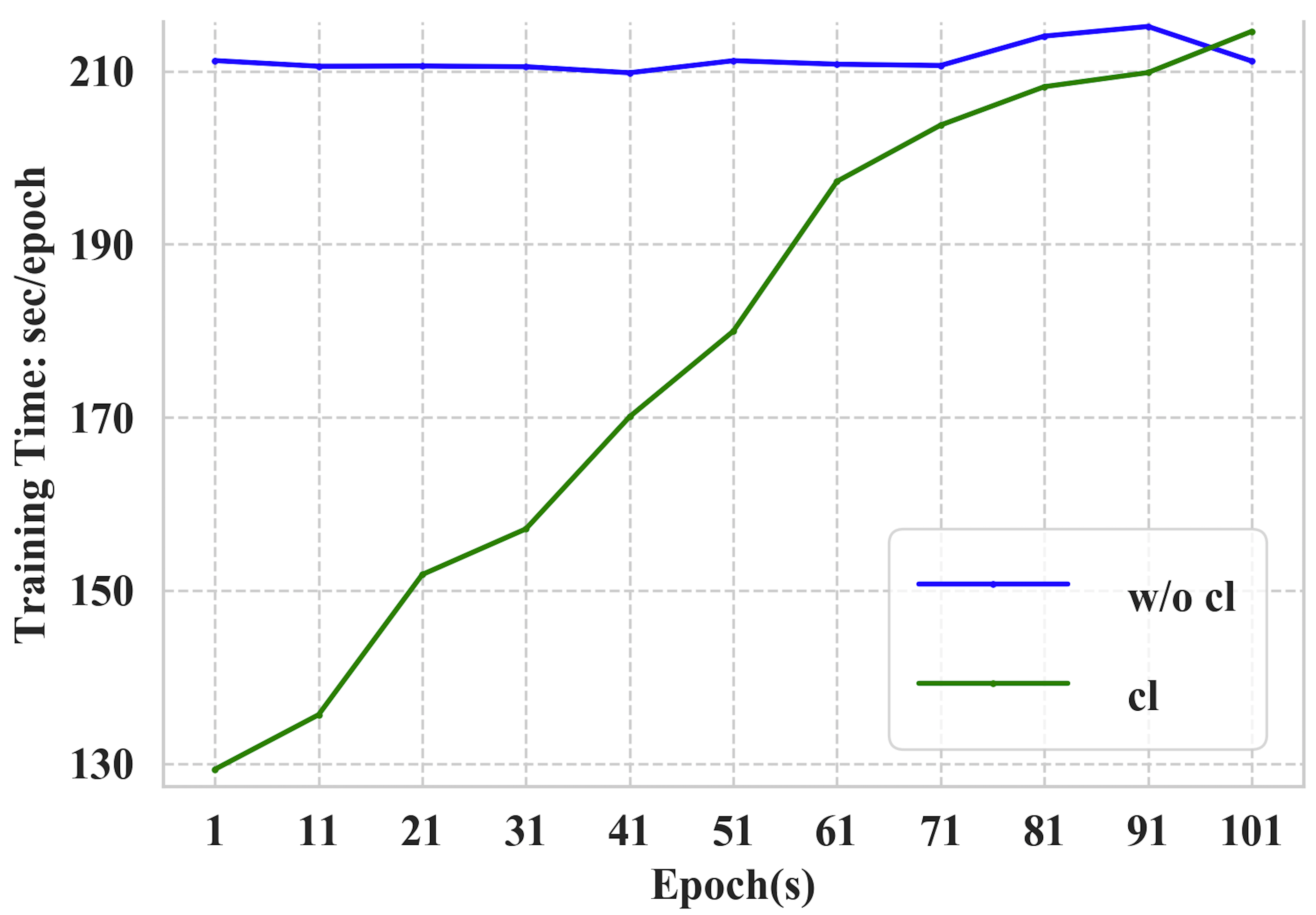}	
	\caption[Efficiency of curriculum learning.]%
    {{\small Efficiency of curriculum learning.}}    
	\label{fig:cl}
\end{figure}

Traditional RNNs' recurrent operation offers flexibility of model architecture, but it brings about time and memory consuming increasing with length of the sequence. Thus, it is necessary to make RNNs perform as few recurrent operations as possible. Meanwhile, if an RNN based decoder is first trained for short-range prediction and has a strong memory for early time steps, it may achieve better performance on longer range after continuous training with longer series. Based on these thoughts and inspired by~\cite{mtgnn}, we employ a general, efficient and effective training strategy for RNN related models, which is called curriculum learning.
During the forward propagation of the training procedure before the back propagation, we do not calculate all steps of the decoder but only first $i$ steps. With the training procedure going on, $i$ keeps increasing until it reaches its maximum, a.k.a, the length of prediction sequence. In this way, it is not necessary for model to keep employing forward and back propagation for all time steps during training, which can significantly decrease the time consuming and also save GPU memory at early stage. More importantly, the performance can also benefit from the strategy because of the good basis for early time steps. The efficiency of curriculum learning on METR-LA dataset can be seen in Figure~\ref{fig:cl}, where the training time can be reduced by as much as 20\% with the performance enhanced simultaneously. We also find the scheduled sampling~\cite{ss} for RNN can work well together with the curriculum learning, which further improve the performance. In our opinion, this training method can be easily adapted to other recurrence based decoder to enhance both efficiency and performance in time series prediction task.

\section{Experiment Settings and Results Analysis}
\subsection{Experiment Settings}
\subsubsection{Datasets}
We conduct experiments on three real-world large-scale datasets:
\begin{figure*}[!t]
	\centering
	\includegraphics[width=0.95\textwidth]{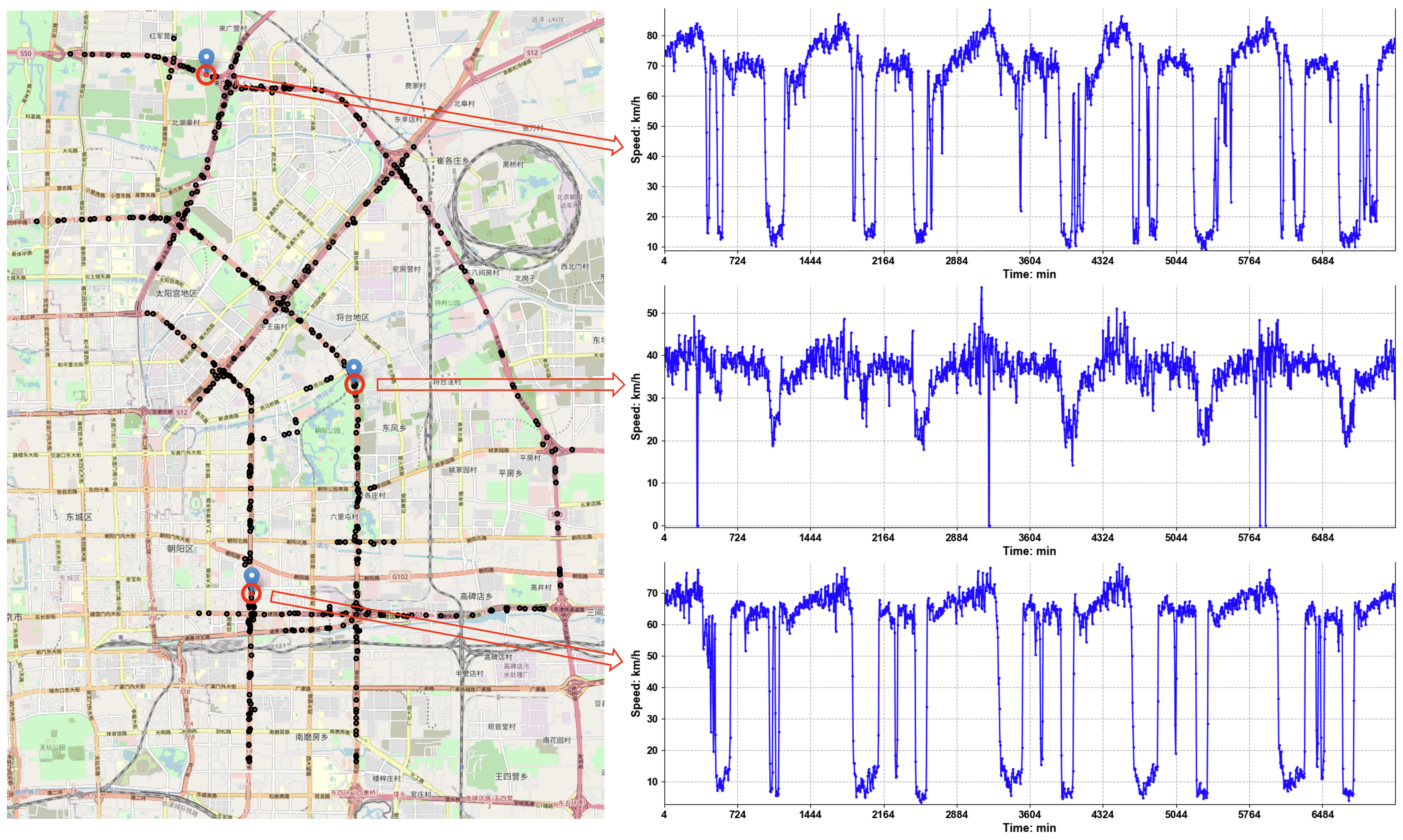}	
	\caption{Road segment distribution of the NE-BJ dataset.}
	\label{fig:NE-BJ500}
\end{figure*}

\begin{itemize}
    \item \textbf{METR-LA} A public traffic speed dataset collected from loop detectors in the highway of Los Angeles containing 207 selected sensors and ranging from Mar 1st 2012 to Jun 30th 2012. The unit of speed is mile/h. In METR-LA, the sensors in different locations are viewed as nodes in the graph.
    
    \item \textbf{PEMS-BAY} A public traffic speed dataset collected by California Transportation Agencies (CalTrans) containing 325 sensors in the Bay Area ranging from Jan 1st 2017 to May 31th 2017. The unit of speed is mile/h. In PEMS-BAY, the sensors in different positions are also considered as nodes in the graph.
    
    \item \textbf{NE-BJ} Previously proposed traffic datasets are either built with vehicle speed on freeways where the traffic speed is relatively high and simple, or fail to fully reflect the complex urban traffic conditions. To overcome these shortcomings and make traffic prediction task more challenging, we publish NE-BJ dataset which is collected and extracted from navigation data of Tencent Map in weekdays of July 2020. The unit of speed is km/h. The dataset contains 500 road segments selected on the main roads in the northeast area of Beijing where a lot more traffic congestion always happens. The NE-BJ dataset is composed of traffic speed data of corresponding road segments seen as nodes in the graph. The distribution of road segments is visualized in Figure~\ref{fig:NE-BJ500}. This dataset can fully reflect the traffic condition of downtown in Beijing and bring out value for further research. 
\end{itemize}

\begin{figure*}[!t]
        \centering
        \begin{subfigure}[b]{0.95\textwidth}   
            \centering 
            \includegraphics[width=\textwidth]{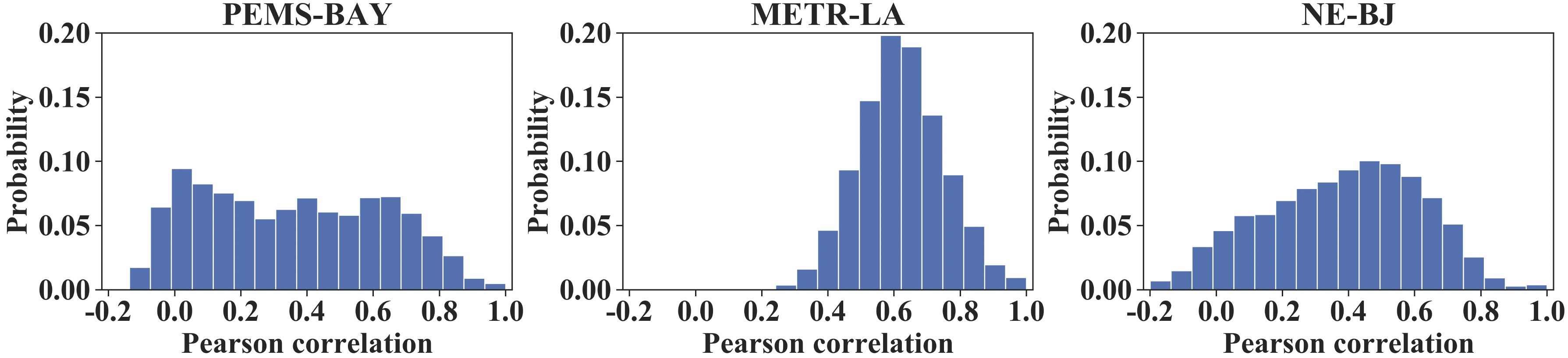}
            \caption[Distributions of inter-node correlations.]%
            {{\small Distributions of inter-node correlations.}}    
            \label{fig:cor}
        \end{subfigure}
        \hfill
        \newline
        \begin{subfigure}[b]{0.95\textwidth}   
            \centering 
            \includegraphics[width=\textwidth]{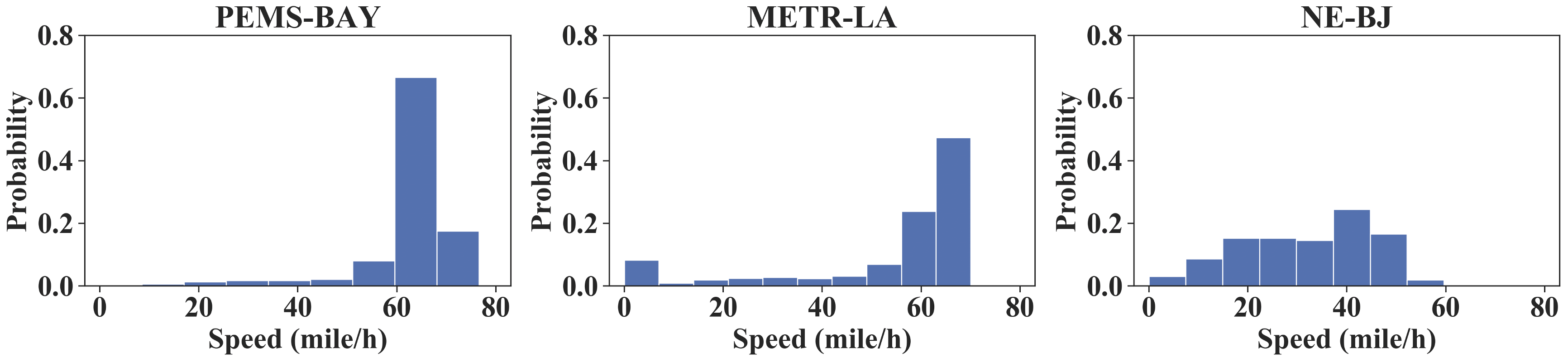}
            \caption[Distributions of speed.]%
            {{\small Distributions of speed.}}    
            \label{fig:dis}
        \end{subfigure}

        \caption[Distributions of inter-node correlations and speed on PEMS-BAY, METR-LA and NE-BJ datasets.]
        {\small Distributions of inter-node correlations and speed on PEMS-BAY, METR-LA and NE-BJ datasets.} 
        \label{fig:para_NE-BJ}
\end{figure*}

To verify it, we conduct statistical analysis on the three datasets, which is shown in Figure~\ref{fig:cor} and Figure~\ref{fig:dis}. In Figure~\ref{fig:cor}, we calculate the Pearson correlations between all pairs of nodes and show the distributions of them. We can observe the explicit spatial correlations in METR-LA while those in PEMS-BAY are distinctly weaker. The internode correlations in NE-BJ dataset is medium and relatively stronger than PEMS-BAY with less low-correlation node pairs. in Figure~\ref{fig:dis}, we display the distributions of speed values under the same unit. The velocity distributions of the PEMS-BAY dataset is obviously much more monotonous and more intensively close to the free-flow velocity, which means simpler traffic conditions and lower correlations between nodes as shown in Figure~\ref{fig:cor}. So it is natural for methods to achieve significant performance on PEMS-BAY dataset. Velocity distributions in METR-LA tend to be polarized with more missing values marked as zero. By comparison, the distributions of traffic speed and spatial correlations in NE-BJ dataset is much more uniform, and completely different from others especially in speed distributions with speed values ranging from 10 mile/h to 50 mile/h, indicating complex traffic conditions and much more congestions. That is why the fine-grained dynamic modeling of internode correlations (such as dynamic adjacency matrix) can significantly improve the performance on NE-BJ dataset with great value for further research.

Following~\cite{dcrnn}, we set the time granularity to 5 minutes and apply Z-Score normalization to the speed data. For METR-LA and PEMS-BAY, 70\% of data is used for training, 20\% are used for testing while the remaining 10\% for validation. For NE-BJ dataset, first 15 days are used for training, last 5 days are used for testing while the remaining 3 days for validation. We see each sensor or road segment as a vertex, where the location of a road segment is represented by its middle point. The pre-defined adjacency matrix is constructed based on pairwise road network distance between sensors (for METR-LA and PEMS-BAY) or the Euclidean distance between road segments (for NE-BJ dataset) with thresholded Gaussian kernel~\cite{6494675}. The pre-defined adjacency matrix $ \mathbf{A} $ can be defined as follows:
\begin{equation}
\mathbf{A}_{v_i, v_j} = \left\{ 
\begin{array}{lr}
\exp ( -\dfrac{d_{v_i,v_j}^2}{\sigma^2} ), if\ d_{v_i,v_j} \leq \kappa \\
0, otherwise 
\end{array}
\label{adjacency_matrix},
\right.
\end{equation}
where $ {d_{v_i,v_j}} $ is the road network distance from sensor $ v_i $ to $ v_j $ or the Euclidean distance between road segment $ v_i $ and $ v_j $. The road network distance is shortest distance the vehicle has to travel from the source to the destination under the constraint imposed by the road network, which is directed. The Euclidean distance is shortest distance on earth which is undirected and calculated only based on the coordinates. $\sigma$ is the standard deviation of distances and $\kappa$ (assigned to 0.1) is the threshold to control the sparsity of $ \mathbf{A} $.
Detailed statistics of the datasets are shown in Table~\ref{tab:data}.

\begin{table}[]
\caption{Statistics of datasets.}
\label{tab:data}
		\resizebox{\columnwidth}{!}{%
\begin{tabular}{cccclclcl}
\hline
Datasets                      &  Samples                  &  Nodes                  & \multicolumn{2}{c}{Sample Rate} & \multicolumn{2}{c}{Input Length} & \multicolumn{2}{c}{Output Length} \\ \hline
NE-BJ                            & 6509                        & 500                      & \multicolumn{2}{c}{5min}        & \multicolumn{2}{c}{12}           & \multicolumn{2}{c}{12}            \\ \hline
METR-LA                       & 34272                       & 207                       & \multicolumn{2}{c}{5min}        & \multicolumn{2}{c}{12}           & \multicolumn{2}{c}{12}            \\ \hline
PEMS-BAY                      & 52116                       & 325                       & \multicolumn{2}{c}{5min}        & \multicolumn{2}{c}{12}           & \multicolumn{2}{c}{12}            \\ \hline
\end{tabular}}
\end{table}

\subsubsection{Parameters Settings}
The proposed model is implemented by Pytorch 1.1.0 on a virtual workstation with a 11G memory Nvidia GeForce RTX 2080Ti GPU. We repeat the experiment 5 times and report the average value of evaluation metrics. The model is trained by the Adam optimizer. The learning rate is set to 0.001. The layer depth of the DGCRM is set to 1 and the size of hidden state is set to 64 for METR-LA and NE-BJ, and 96 for PEMS-BAY. $\alpha,\beta,\gamma$ are set to 0.05, 0.95 and 0.95, respectively. 
The dimension of node embeddings is set to 40 for METR-LA and PEMS-BAY, and 100 for NE-BJ. 
The batch size is set to 64 on METR-LA and PEMS-BAY, and 16 for NE-BJ.
Early stopping is employed to avoid overfitting. We use MAE as our model's loss function for training.

\subsubsection{Baselines}
We compare DGCRN
with traditional statistic-based methods and recently proposed GNN based models for traffic prediction task, which can fully reflect the latest progress in this field. The baselines are introduced as follows:

\begin{itemize}
    \item \textbf{HA:} Historical Average (HA) method is used to predict the future speed using the average value of historical speed data.
    \item \textbf{VAR~\cite{hamilton1994time}:} Vector Auto-Regression (VAR) can be used for time series forecasting.
    \item \textbf{SVR:} Support Vector Regression (SVR) is another classical time series analysis model which uses linear support vector machine for the regression task.
    \item \textbf{FNN:} This is a feed forward neural network for time series prediction.
    \item \textbf{ARIMA$_{kal}$~\cite{box1970time}:} This is a traditional and widely used method in time series prediction, which integrates auto-regression with moving average model.
    \item \textbf{FC-LSTM~\cite{lstm}:} Long Short Term Memory (LSTM) network with fully connected hidden units is a well-known network architecture which is powerful in capturing sequential dependency.
    \item \textbf{DCRNN~\cite{dcrnn}:} This is a GNN-based and RNN-based model which integrates GRU with dual directional diffusion convolution. 
    \item \textbf{STGCN~\cite{stgcn}:} It is a GNN-based and CNN-based model which incorporates graph convolutions with 1D convolutions
    \item \textbf{Graph WaveNet~\cite{gwn}:} It is a GNN-based and CNN-based model which integrates diffusion graph convolutions with gated 1D dilated convolutions and propose self-adaptive adjacency matrix.
    \item \textbf{ST-MetaNet~\cite{stmetanet}:} It is a GNN-based and RNN-based model which employs meta learning to generate parameters of GAT and GRU.
    \item \textbf{AGCRN~\cite{agcrn}:} This is a GNN-based and RNN-based model which employs adaptive graph and integrates GRU with graph convolutions with node adaptive parameter learning.
    \item \textbf{GMAN~\cite{gman}:} This is an attention-based model with spatial, temporal and transform attentions.
    \item \textbf{MTGNN~\cite{mtgnn}:} It is a GNN-based and CNN-based model which employs adaptive graph, mix-hop propagation layers and dilated inception layers to capture spatio-temporal correlations.

\end{itemize}

For all the baselines, we use the default settings from their original proposals. The performances of all methods are evaluated by three  commonly  used  metrics in traffic prediction, including (1) Mean Absolute Error (MAE), which is a basic metric to reflect the actual situation of the prediction accuracy. (2) Root Mean Squared Error (RMSE), which is more sensitive to abnormal value, and (3) Mean  Absolute  Percentage  Error (MAPE) which can eliminate the influence of data unit to some extent, defined as follows: 

\begin{equation}
\begin{split}
\label{eq:metrics}
&\text{MAE}( {x}, \hat{ {x}}) = \frac{1}{| {\Omega}|} \sum_{i \in  {\Omega}} \left|x_i - \hat{x}_i\right|,\\
&\text{RMSE}( {x}, \hat{ {x}}) = \sqrt{\frac{1}{| {\Omega}|} \sum_{i \in  {\Omega}} (x_i - \hat{x}_i)^2},\\
&\text{MAPE}( {x}, \hat{ {x}}) = \frac{1}{| {\Omega}|} \sum_{i \in  {\Omega}} \left|\frac{x_i - \hat{x}_i}{x_i}\right|,\\
\end{split}
\end{equation}
where $ {x}=x_1, \cdots, x_n$ denotes the ground truth, $\hat{ {x}}=\hat{x}_1, \cdots, \hat{x}_n$ represents the predicted values, and $\Omega$ denotes the indices of observed samples. In our experiments, the $|  {\Omega} |$ is 12.

\subsection{Traffic Benchmark}
\subsubsection{Background}
With rapid growing of traffic prediction field, more and more models have been proposed with corresponding experiments such as ablation studies, parameter studies and case studies. However, the models are evaluated on different datasets with various experimental settings, making it tricky for researchers to make comparisons, reproduce works or develop their own models. Thus, to build a public platform where related works are evaluated on same datasets and metrics, and to enhance the reproducibility, we publish an open-source traffic benchmark where it is easy to reproduce all kinds of models on various datasets. The comparisons among models in the benchmark are shown in Table~\ref{table:benchmark}. 
Specifically, the benchmark is composed of the performance comparisons for 15 minutes (horizon 3), 30 minutes (horizon 6) and 1 hour (horizon 12) ahead forecasting on three datasets.
all the missing values in the test datasets are masked and excluded when calculating metrics and loss. We employ data completion on the missing values of the input sequences in NE-BJ dataset with the speed value of last time step.

\subsubsection{Results}
\begin{table*}[!h]
\scriptsize
	\centering
	\caption{Benchmark for traffic speed prediction task.} 
	\label{table:benchmark}
	\resizebox{1\textwidth}{!}{
		
		\begin{tabular}{ l  c c r | c c r |c c r}
			\toprule
			\hline
			\multirow{2}{*}{}   & \multicolumn{3}{c}{Horizon 3} & \multicolumn{3}{c}{Horizon 6}  & \multicolumn{3}{c}{Horizon 12} \\
			\cline{2-4}  \cline{5-7} \cline{8-10}   		    \\[-1em]  
			
			& {\footnotesize MAE} & {\footnotesize RMSE} & {\footnotesize MAPE} & {\footnotesize MAE} & {\footnotesize RMSE} & {\footnotesize MAPE} & {\footnotesize MAE} & {\footnotesize RMSE} & {\footnotesize MAPE}\\
			\midrule
			\hline
			METR-LA  \\ \cline{1-1}
			\\[-1em]  
			
			HA & 4.16 & 7.80 & 13.00\% & 4.16 & 7.80  & 13.00\% & 4.16 & 7.80  & 13.00\% \\
            ARIMA$_{kal}$ & 3.99 & 8.21 & 9.60\%  & 5.15 & 10.45 & 12.70\% & 6.90 & 13.23 & 17.40\% \\
            VAR & 4.42 & 7.89 & 10.20\% & 5.41 & 9.13  & 12.70\% & 6.52 & 10.11 & 15.80\% \\
            SVR & 3.99 & 8.45 & 9.30\%  & 5.05 & 10.87 & 12.10\% & 6.72 & 13.76 & 16.70\% \\
            FNN & 3.99 & 7.94 & 9.90\%  & 4.23 & 8.17  & 12.90\% & 4.49 & 8.69  & 14.00\% \\
            FC-LSTM & 3.44 & 6.30 & 9.60\%  & 3.77 & 7.23  & 10.90\% & 4.37 & 8.69  & 13.20\% \\
			DCRNN  & 2.77 & 5.38 & 7.30\% & 3.15 & 6.45 & 8.80\% & 3.60 & 7.60 & 10.50\% \\ 
			STGCN  & 2.88 & 5.74 & 7.62\% & 3.47 & 7.24 & 9.57\% & 4.59 & 9.40 & 12.70\%\\
			Graph WaveNet  & 2.69 & 5.15 & 6.90\% & 3.07 & 6.22 & 8.37\% & 3.53 & 7.37 & 10.01\%\\
			ST-MetaNet & 2.69 & 5.17 & 6.91\% & 3.10 & 6.28 & 8.57\% & 3.59 & 7.52 & 10.63\% \\
            ASTGCN & 4.86 & 9.27 & 9.21\%  & 5.43 & 10.61 & 10.13\% & 6.51 & 12.52 & 11.64\% \\
            STSGCN & 3.31 & 7.62 & 8.06\%  & 4.13 & 9.77  & 10.29\%  & 5.06 & 11.66  & 12.91\% \\
            AGCRN & 2.87 & 5.58 & 7.70\%  & 3.23 & 6.58  & 9.00\%  & 3.62 & 7.51  & 10.38\% \\
			GMAN & 2.80 & 5.55 & 7.41\%  & 3.12 & 6.49  & 8.73\%  & 3.44 & 7.35  & 10.07\% \\	    
			MTGNN & 2.69 &  5.18 & 6.86\% & 3.05 & 6.17 & 8.19\% & 3.49 & 7.23 & 9.87\%\\
			\\[-1em]  
			\hline
			\\[-1em]

             DGCRN & \textbf{2.62} & \textbf{5.01} & \textbf{6.63\%}  & \textbf{2.99} & \textbf{6.05}  & \textbf{8.02\%}  & \textbf{3.44} & \textbf{7.19}  & \textbf{9.73\%} \\
			\hline \hline 
			PEMS-BAY  \\ \cline{1-1}
			\\[-1em] 
			HA & 2.88 & 5.59 & 6.80\% & 2.88 & 5.59 & 6.80\% & 2.88 & 5.59 & 6.80\%  \\
            ARIMA$_{kal}$ & 1.62 & 3.30 & 3.50\% & 2.33 & 4.76 & 5.40\% & 3.38 & 6.50 & 8.30\%  \\
            VAR & 1.74 & 3.16 & 3.60\% & 2.32 & 4.25 & 5.00\% & 2.93 & 5.44 & 6.50\%  \\
            SVR & 1.85 & 3.59 & 3.80\% & 2.48 & 5.18 & 5.50\% & 3.28 & 7.08 & 8.00\%  \\
            FNN & 2.20 & 4.42 & 5.19\% & 2.30 & 4.63 & 5.43\% & 2.46 & 4.98 & 5.89\%  \\
            FC-LSTM & 2.05 & 4.19 & 4.80\% & 2.20 & 4.55 & 5.20\% & 2.37 & 4.96 & 5.70\%  \\
			DCRNN  & 1.38 & 2.95 & 2.90\% & 1.74 & 3.97 & 3.90\% & 2.07 & 4.74 & 4.90\% \\ 
			STGCN  & 1.36 & 2.96 & 2.90\% & 1.81 & 4.27 & 4.17\% & 2.49 & 5.69 & 5.79\%\\
			Graph WaveNet  & 1.30 & 2.74 & 2.73\% & 1.63 & 3.70& 3.67\% & 1.95 &  4.52 &  4.63\%\\
			ST-MetaNet & 1.36 & 2.90 & 2.82\% & 1.76 & 4.02 & 4.00\% & 2.20 & 5.06 & 5.45\%\\
            ASTGCN & 1.52 & 3.13 & 3.22\% & 2.01 & 4.27 & 4.48\% & 2.61 & 5.42 & 6.00\%  \\
            STSGCN & 1.44 & 3.01 & 3.04\% & 1.83 & 4.18 & 4.17\% & 2.26 & 5.21 & 5.40\% \\
            AGCRN & 1.37 & 2.87 & 2.94\% & 1.69 & 3.85 & 3.87\% & 1.96 & 4.54 & 4.64\%  \\
			GMAN  & 1.34 & 2.91 & 2.86\% & 1.63 & 3.76 & 3.68\% & \textbf{1.86} & \textbf{4.32} & \textbf{4.37\%}  \\		 
			MTGNN & 1.32 & 2.79& 2.77\% & 1.65& 3.74 & 3.69\% & 1.94 & 4.49 & 4.53\%\\
            \\[-1em]  
			\hline
			\\[-1em]  
            DGCRN & \textbf{1.28} & \textbf{2.69} & \textbf{2.66\%} & \textbf{1.59} & \textbf{3.63} & \textbf{3.55\%} & 1.89 & 4.42 & 4.43\%  \\

            \hline \hline 
			NE-BJ  \\ \cline{1-1}
			\\[-1em] 
			HA & 6.00 & 10.95 & 26.40\% & 6.00 & 10.95 & 26.40\% & 6.00 & 10.95 & 26.40\% \\
            VAR & 5.42 & 8.16  & 19.28\% & 5.76 & 9.07  & 21.53\% & 6.14 & 9.65  & 23.33\% \\
            FNN & 4.08 & 7.22  & 13.31\% & 5.14 & 9.27  & 18.24\% & 6.47 & 11.35 & 25.57\% \\
            FC-LSTM & 3.97 & 7.05  & 13.05\% & 4.93 & 9.04  & 17.74\% & 6.06 & 10.88 & 23.52\% \\
			DCRNN  & 3.84 & 6.84  & 12.82\% & 4.51 & 8.49  & 15.84\% & 5.15 & 9.77  & 19.08\% \\
			STGCN  & 5.02 & 8.34  & 19.31\% & 5.10 & 8.55  & 19.82\% & 5.39 & 9.09  & 22.14\% \\
			Graph WaveNet  & 3.74 & 6.54  & 12.49\% & 4.41 & 8.08  & 15.79\% & 4.99 & 9.20  & 19.45\% \\
			ST-MetaNet & 3.82 & 6.69  & 13.05\% & 4.50 & 8.46  & 16.93\% & 5.05 & 9.74  & 20.00\% \\
			ASTGCN & 4.43 & 7.34 & 14.65\% & 5.31 & 8.86 & 18.24\% & 6.29 & 10.31 & 22.70\% \\
            AGCRN & 3.84 & 6.75  & 13.80\% & 4.48 & 8.41  & 16.70\% & 4.99 & 9.44  & 19.94\% \\
			GMAN  & 4.08 & 7.63  & 14.94\% & 4.42 & 8.45  & 16.51\% & 4.80 & \textbf{9.18}  & 18.36\% \\	 
			MTGNN & 3.75 & 6.71  & 12.91\% & 4.39 & 8.33  & 16.07\% & 4.90 & 9.38  & 19.79\% \\
            \\[-1em]  
			\hline
			\\[-1em]  
            DGCRN & \textbf{3.56} & \textbf{6.27}  & \textbf{12.01\%} & \textbf{4.23} & \textbf{7.96}  & \textbf{15.10\%} & \textbf{4.79} & 9.23  & \textbf{17.98\%} \\
            \hline
			
			\bottomrule
		\end{tabular}
	}
\end{table*}

Table~\ref{table:benchmark} shows the performances of baselines and our model. DGCRN achieves state-of-the-art prediction performances on the three datasets (2\%$\sim$4\% improvements on METR-LA and PEMS-BAY datasets) and the advantages are more distinct for complex traffic conditions (around 6\% on NE-BJ dataset). DGCRN can generally outperform other baselines for most forecasting steps except for certain metrics in the long-range horizon (e.g., 1 hour ahead) in PEMS-BAY and NE-BJ compared with GMAN, which suggests the effectiveness of dynamic graph modeling. We employ a hyper-network to generate filter which can dynamically adjust the weights of graph based on temporal information, which is critical for dynamic graph generation and performance improvement. What's more, traditional time series methods perform much worse than deep learning methods while the graph-based methods further enhance the performance significantly, indicating the power of deep learning as well as the effectiveness of road network information.
GMAN employs self-attention based architectures and spatio-temporal embedding module which is good for long-range prediction. However, self-attention can not capture local sequential correlations and spatio-temporal embedding is relatively simple for modeling the complex spatio-temporal dependency, which degrade the performance for short-range prediction. Graph WaveNet, MTGNN and AGCRN benefit a lot from the self-adaptive adjacency matrix whose parameters are learnable and can be updated with back propagation algorithm. However, the adaptive graph is still static with time changes and fails to capture dynamic spatial dependencies at each time step. Besides, most of the other baselines fail to model the dynamic characteristics of the traffic network structure, which restricts the representation ability. ASTGCN and STSGCN perform not well on the three datasets, which is possibly because of the missing values and the restricted representation ability of the models.

We can also find that difficulty of traffic prediction task varies significantly with different traffic datasets. The NE-BJ dataset is much more complicated than others, leading to worse performance for all models. The prediction task is easier on PEMS-BAY making the performace much better obviously. Besides, ARIMA and STSGCN break down when employed on the NE-BJ dataset because of the complexity of traffic speed series, so we do not list them in the benchmark. 

\subsubsection{Ablation Study}
\begin{table*}[!h]
\scriptsize
	\centering
	\caption{Ablation study on METR-LA.} 
	\label{table:ablation}
	\resizebox{1\textwidth}{!}{
		
        \begin{tabular}{ l  c c r | c c r |c c r}
			\toprule
			\hline
			\multirow{2}{*}{}   & \multicolumn{3}{c}{Horizon 3} & \multicolumn{3}{c}{Horizon 6}  & \multicolumn{3}{c}{Horizon 12} \\
			\cline{2-4}  \cline{5-7} \cline{8-10}   		    \\[-1em]  
			
			& {\footnotesize MAE} & {\footnotesize RMSE} & {\footnotesize MAPE} & {\footnotesize MAE} & {\footnotesize RMSE} & {\footnotesize MAPE} & {\footnotesize MAE} & {\footnotesize RMSE} & {\footnotesize MAPE}\\
			\midrule
			\hline
			\\[-1em]  
            DGCRN & \textbf{2.62} & \textbf{5.01} & \textbf{6.63\%}  & \textbf{2.99} & \textbf{6.05}  & \textbf{8.02\%}  & \textbf{3.44} & \textbf{7.19}  & \textbf{9.73\%} \\
			
			w/o dg            & 2.71 & 5.19 & 7.04 \% & 3.12 & 6.31 & 8.60 \% & 3.64 & 7.59 & 10.62 \% \\
			w/o preA            & 2.64 & 5.09 & 6.68 \% & 3.02 & 6.15 & 8.14 \% & 3.47 & 7.30 & 9.97 \% \\
			w/o hypernet    & 2.67 & 5.12 & 6.90 \% & 3.05 & 6.20 & 8.33 \% & 3.50 & 7.39 & 10.03 \% \\
			dg w/o speed      & 2.63 & 5.01 & 6.67 \% & 3.02 & 6.08 & 8.10 \% & 3.49 & 7.29 & 9.90 \%  \\
			dg w/o time       & 2.62 & 5.03 & 6.65 \% & 3.00 & 6.09 & 8.08 \% & 3.46 & 7.29 & 9.83 \%  \\
			dg w/o h          & 2.63 & 5.04 & 6.72 \% & 3.00 & 6.08 & 8.23 \% & 3.45 & 7.23 & 10.04 \% \\
			dg2sg           & 2.65 & 5.05 & 6.73 \% & 3.02 & 6.10 & 8.17 \% & 3.48 & 7.28 & 9.96 \%  \\

			w/o cl              & 2.63 & 5.04 & 6.67 \% & 3.00 & 6.13 & 8.07 \% & 3.47 & 7.35 & 9.77 \% \\
			
			hypernet mul2matmul & 2.67 & 5.12 & 6.86 \% & 3.05 & 6.19 & 8.30 \% & 3.50 & 7.36 & 10.06 \% \\
            \hline
			
			\bottomrule
		\end{tabular}
	}
\end{table*}

To evaluate the effect of key components that contribute to the improved outcomes of our proposed model, we conduct an ablation study on the METR-LA dataset. We name the variants of DGCRN as follows:

\begin{itemize}
	\item \textbf{w/o dg}: This is DGCRN without dynamic adjacency matrix. We remove the dynamic graph convolution from DGCRM.
	\item \textbf{w/o preA}: We remove the pre-defined graph convolution from DGCRM.
	\item \textbf{w/o hypernet}: We replace the hyper-networks in the dynamic graph generator with a simple fully connected layer.

    \item \textbf{dg w/o time}: This is DGCRN whose dynamic graph is generated without time of day as inputs. 
    \item \textbf{dg w/o speed}: This is DGCRN whose dynamic graph is generated without speed as inputs.
	\item \textbf{dg w/o h}: This is DGCRN whose dynamic graph is generated without hidden state as inputs. 
	\item \textbf{dg2sg}: This is DGCRN whose graph is not updated step by step, we call it static graph (sg). The generation method of sg is as same as MTGNN~\cite{mtgnn}.
	\item \textbf{w/o cl}: This is DGCRN without curriculum learning. We train DGCRN only with scheduled sampling.
	\item \textbf{hypernet mul2matmul}: We replace the Hadamard product in equations~\ref{eq:dgg} with matrix multiplication.

\end{itemize}

We repeat each experiment 5 times with early-stopping per repetition and report the average of MAE, RMSE, MAPE on the test set in Table \ref{table:ablation}. The introduction of dynamic graph significantly improves the performance as it provides an self-adaptive adjacency matrix generated based on dynamic information at each step, thus can collaborate with static pre-defined adjacency matrix to describe the dynamic topology of road network in a better way. The proposed training strategy is evidently effective (\textbf{w/o cl}). The ablation study on the inputs of dynamic graph generator, i.e., the deficiency of hidden state (\textbf{dg w/o h}), time of day (\textbf{dg w/o time}), speed (\textbf{dg w/o speed}) or dynamic characteristic (\textbf{dg2sg}) will degrade the performance, especially for long-range prediction. 
\textbf{hypernet mul2matmul} requires reshaping the dynamic filter tensor to $\mathbf{R}^{B\times N\times D_{in}\times D_{out}}$ for matrix multiplication. We can see the sharp decrease of the performance with the parameters of dynamic filters 40 times as much as those before.

In summary, both dynamic graph and static graph are essential for the performance of DGCRN. The hyper-networks confirm the effectiveness of the dynamic graph generation and all the inputs of dynamic graph generator can help the model perform better. DGCRN captures both static and dynamic spatio-temporal dependence to achieve good results.

\subsubsection{Parameter Study}

\begin{figure}[!h]
        \begin{subfigure}[!h]{0.24\textwidth}   
            \centering 
            \includegraphics[width=\textwidth]{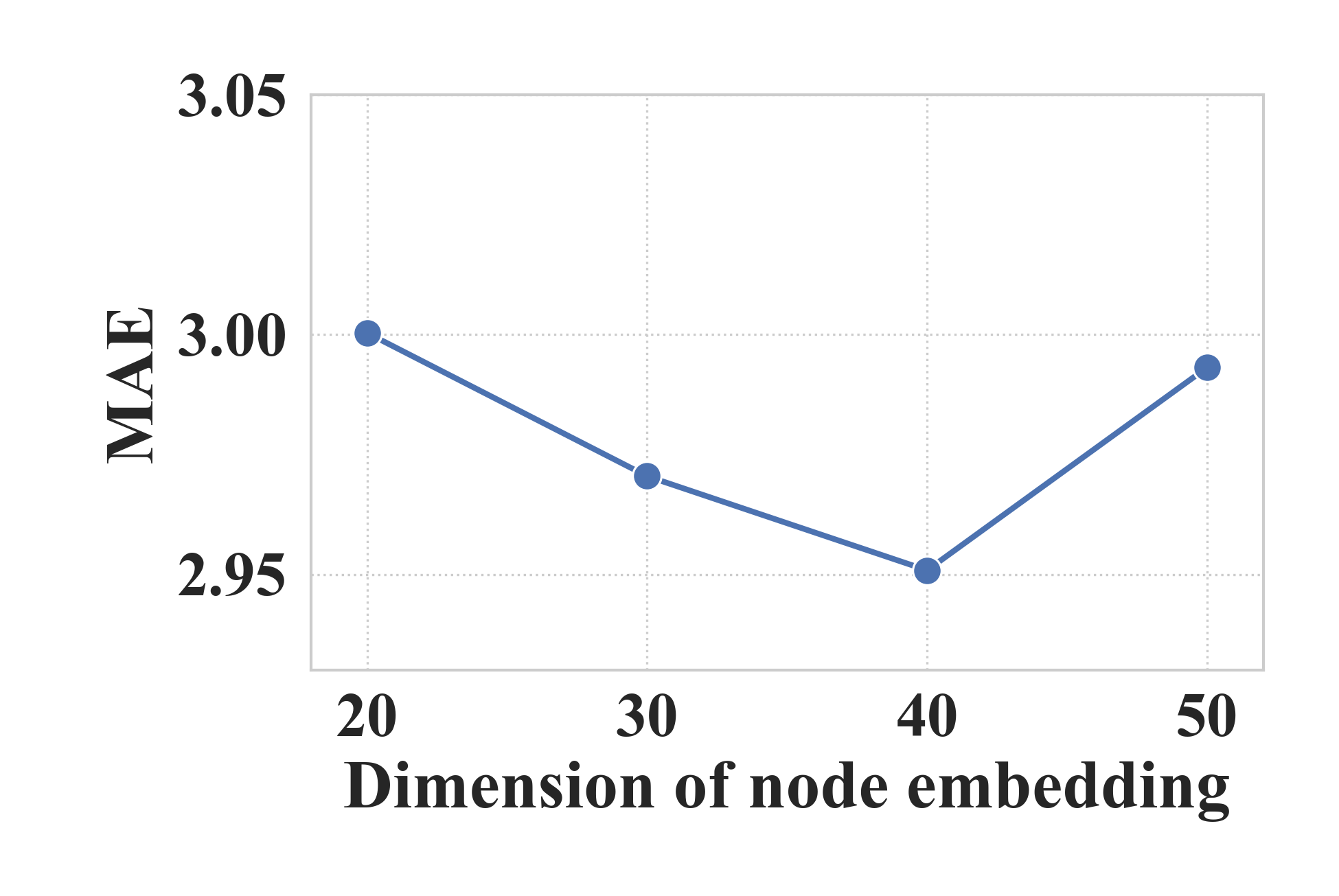}
            \caption[Effects of dimension of node embedding.]%
            {{\small Effects of dimension of node embedding.}}    
            \label{fig:pa}
        \end{subfigure}
        \hfill
        \begin{subfigure}[!h]{0.24\textwidth}   
            \centering 
            \includegraphics[width=\textwidth]{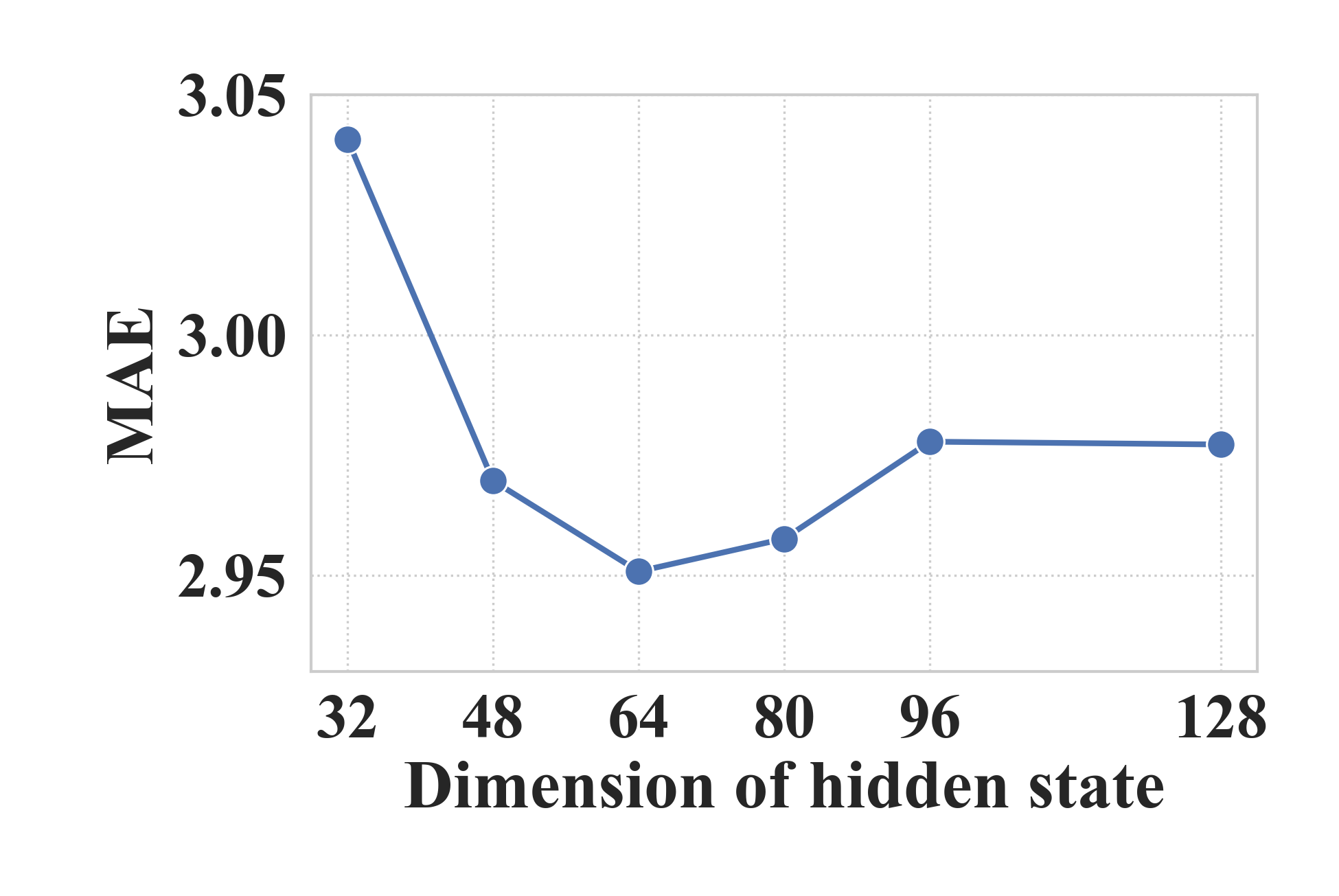}
            \caption[Effects of dimension of hidden state.]%
            {{\small Effects of dimension of hidden state.}}    
            \label{fig:pb}
        \end{subfigure}
        \newline
        
        \begin{subfigure}[!h]{0.24\textwidth} 
        	\centering
        	\includegraphics[width=1.00\textwidth]{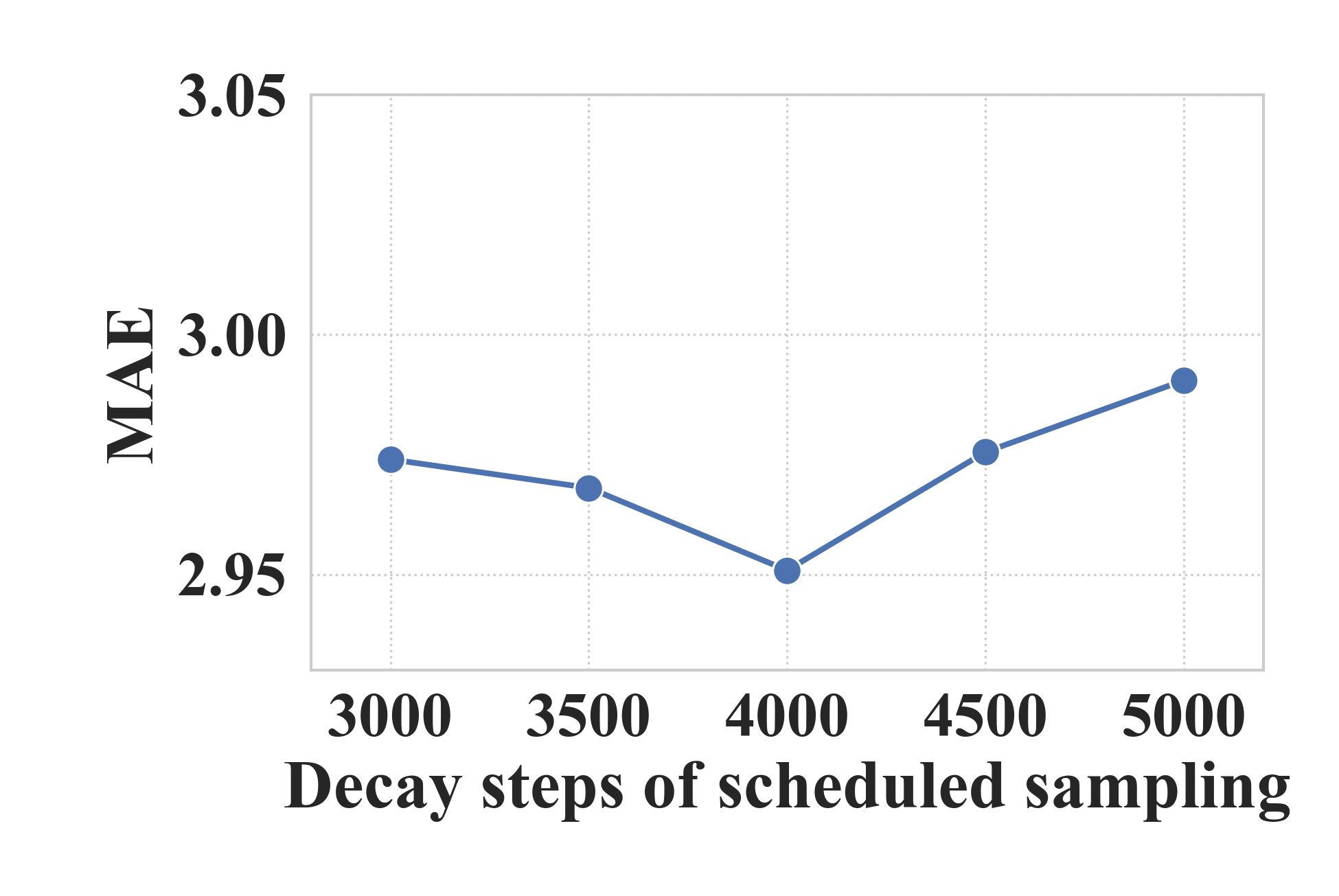}	
        	\caption[Effects of decay steps of scheduled sampling.]%
            {{\small Effects of decay steps of scheduled sampling.}}    
        	\label{fig:ss}
        \end{subfigure}
        \hfill
        \begin{subfigure}[!h]{0.24\textwidth}   
            \centering 
            \includegraphics[width=\textwidth]{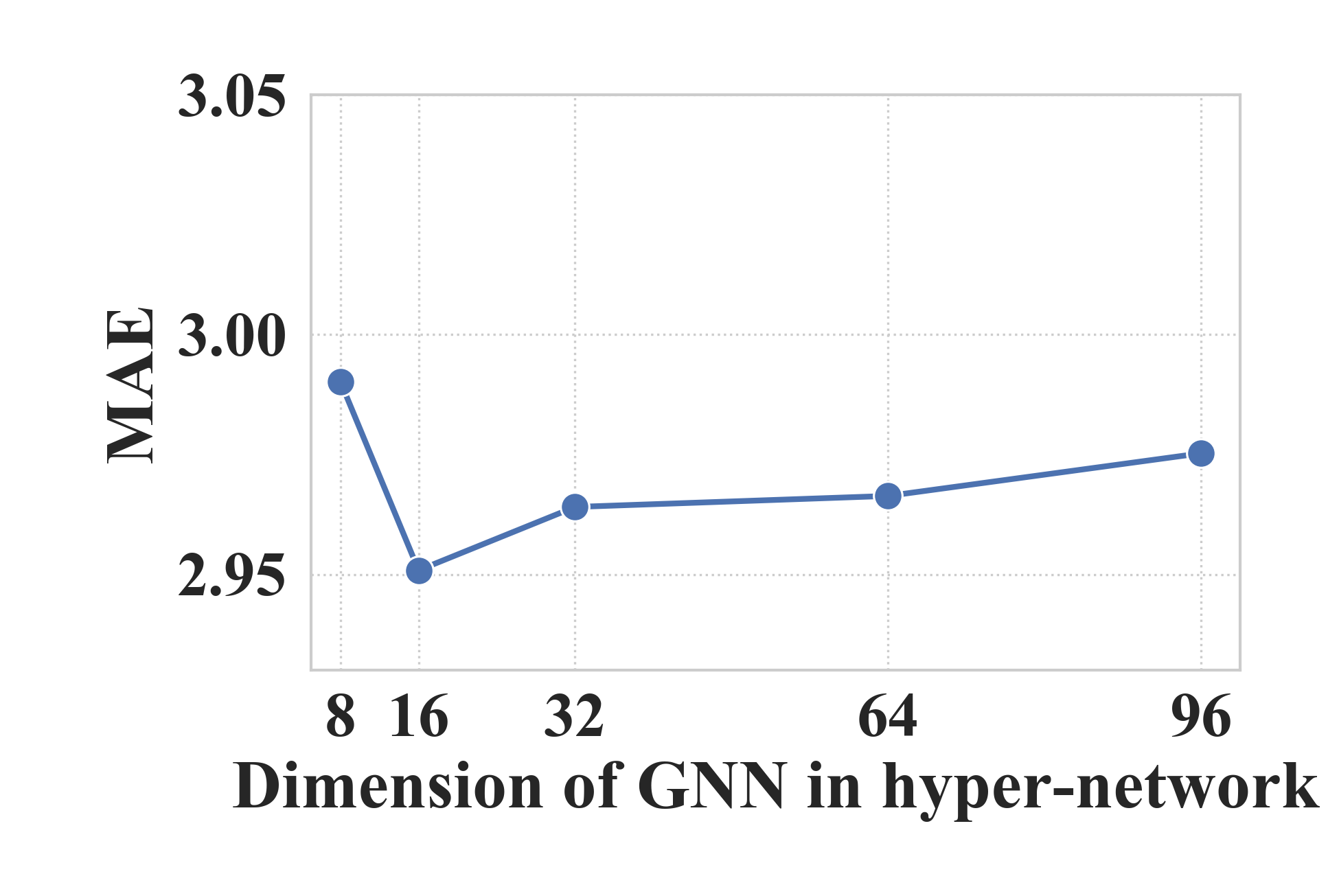}
            \caption[Effects of output dimension of GNN in hyper-network.]%
            {{\small Effects of output dimension of GNN in hyper-network.}}    
            \label{fig:hyperGNN_dim}
        \end{subfigure}
        \caption[Parameter Study on METR-LA.]
        {\small Parameter Study on METR-LA.} 
        \label{fig:para}
\end{figure}

\begin{figure}[!h]
        \begin{subfigure}[!]{0.24\textwidth}   
            \centering 
            \includegraphics[width=\textwidth]{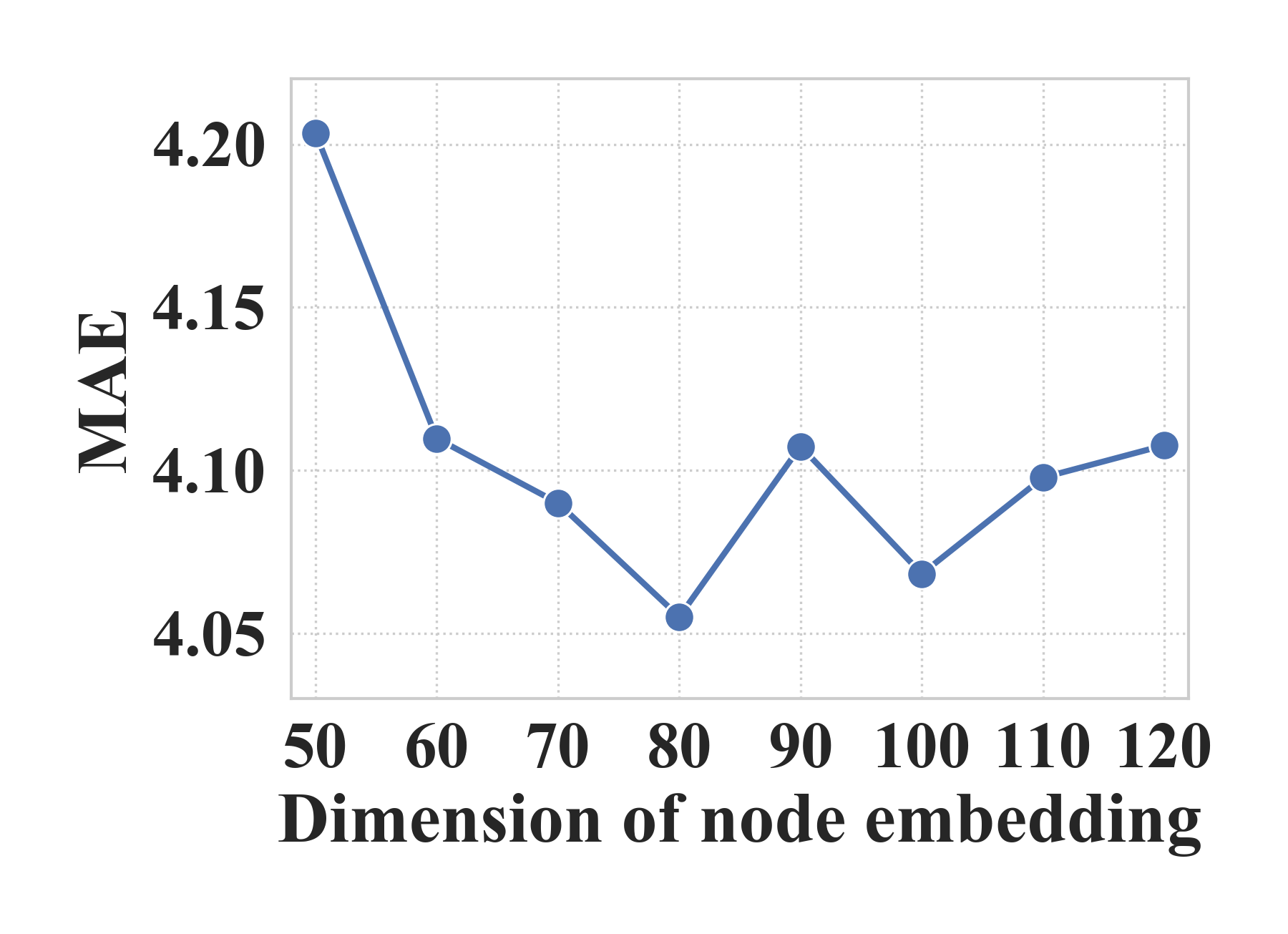}
            \caption[Effects of dimension of node embedding.]%
            {{\small Effects of dimension of node embedding.}}    
            \label{fig:pa_NE-BJ}
        \end{subfigure}
        \hfill
        \begin{subfigure}[!]{0.24\textwidth}   
            \centering 
            \includegraphics[width=\textwidth]{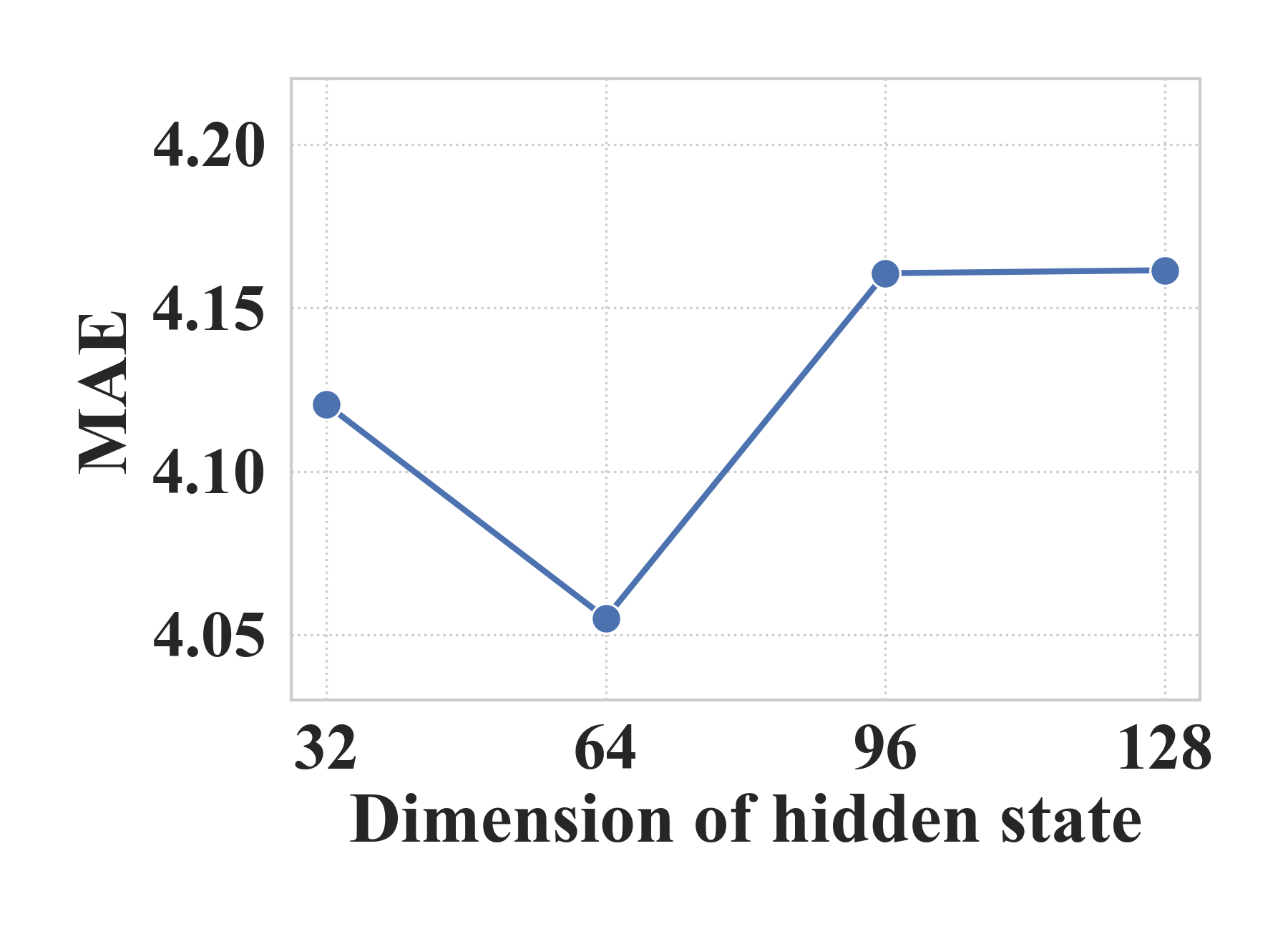}
            \caption[Effects of dimension of hidden state.]%
            {{\small Effects of dimension of hidden state.}}    
            \label{fig:pb_NE-BJ}
        \end{subfigure}

        \caption[Parameter Study on NE-BJ.]
        {\small Parameter Study on NE-BJ.} 
        \label{fig:para_NE-BJ}
\end{figure}
To study the influence of hyper-parameters, we conduct a parameter study on core hyper-parameters of DGCRN. The chosen hyper-parameters are as follows: Decay steps of scheduled sampling, ranges from 3000 to 5000. Dimension of node embedding, the dimension of the dynamic node embeddings, ranges from 20 to 60 on METR-LA and from 50 to 120 on NE-BJ. Dimension of hidden state, the dimension of RNN's hidden state in DGCRN, ranges from 32 to 128. 
Dimension of GNN in hyper-network, the output dimension of GNN, ranges from 8 to 96.

We repeat each experiment 5 times and report the average of MAE on the test set. In each experiment, we only change one parameter while fixing all other parameters. Figure~\ref{fig:para} and Figure~\ref{fig:para_NE-BJ} show the experimental results of parameter study. As shown in Figure~\ref{fig:pa}, Figure~\ref{fig:pb} and Figure~\ref{fig:pb_NE-BJ}, increasing the dimension of both the dynamic node embeddings and the hidden state will improve the representation ability of DGCRN, while reducing the MAE loss. However, the dimension of hidden state of DGCRM should not be too high or it will result in over-fitting, degrading the performance of model obviously. There is an optimal value for the dimension of hidden state at around 64. On METR-LA, the model is not sensitive to a high dimension of node embeddings comparing with the dimension of hidden state and it is good for model when the dimension of node embeddings is 40. However, on NE-BJ dataset, the performance of DGCRN can achieve outstanding performance significantly with a high dimension of node embeddings at around 80 or 100, as illustrated in Figure~\ref{fig:pa_NE-BJ}. This is because the traffic conditions of NE-BJ dataset is much more complex, requiring node embeddings to have higher dimension with the ability to store enough knowledge. In addition, the decay steps of scheduled sampling is best at around 4000 on METR-LA, as shown in Figure~\ref{fig:ss}. Figure~\ref{fig:hyperGNN_dim} demonstrates the effect of the dimension of GNN in hyper-network. When the dimension of GNN in hyper-network changes from 8 to 16, the model gets an obvious benefit, which indicates that the dynamic graph generation is effective in capturing dynamic spatio-temporal correlations. However, the dimension of GNN in hyper-network should not be too high in case of over-fitting.

To sum up, all the hyper-parameters have optimal values and should be confirmed based on the type of datasets, which will influence the ultimate performance of model to a large extent.

\section{Conclusion}
In this paper, we take the first step to explore the generation and application of step-wise dynamic graph in traffic prediction, preliminarily proving that dynamic graph can effectively cooperate with pre-defined graph while improving the prediction performance. Based on dynamic graph generation, we propose DGCRN for traffic prediction.
As core of model, the dynamic graph is able to learn the parameters of adjacency matrix, and make self-adjustment according to the dynamic information at each time step. 
Besides, we employ a general training method for RNNs, which is proven more efficient and good for performance.
Moreover, the NE-BJ dataset is representative for urban traffic conditions with great value for further research. In addition, we provide the traffic benchmark for researchers to conduct fair comparisons of different methods and design their own models conveniently. Finally, we conduct sufficient experiments to prove the dynamic graph generation is promising. As our future work, we plan to go deeper into the urban traffic prediction by dealing with traffic congestion. Additionally, more external factors (e.g., temperature, weather, POI, accidents, etc.) can be integrated into the model to further improve the performance.

\bibliographystyle{IEEEtran}
\bibliography{reference.bib}
\end{document}